\RequirePackage{fix-cm}
\documentclass[twocolumn]{svjour3}          % twocolumn
\smartqed  % flush right qed marks, e.g. at end of proof
\usepackage{graphicx}
\usepackage{amsmath}
\usepackage{multirow}
\usepackage{pifont}
\usepackage{graphicx}
\usepackage{amsmath}
\usepackage{amssymb}
\usepackage{array}
\usepackage{multirow, nicefrac}
\usepackage{pifont}
\usepackage{cite}
\usepackage{bbm}
\usepackage{arydshln}
\usepackage{xcolor}
\usepackage{tablefootnote}
\usepackage{threeparttable}
\definecolor{green}{rgb}{1,0,0}

\def\eg{\emph{e.g.}}
\def\ie{\emph{i.e.}}

\def\etal{\emph{et al.}}

\def\etal{{\em et al.\/}\, }

\def\btheta{\mbox{{\boldmath $\theta$}}}

\def\0{{\bf 0}}
\def\1{{\bf 1}}

\def\ba{{\bf a}}

\def\bg{{\bf g}}

\def\bw{{\bf w}}
\def\bx{{\bf x}}

\def\st{\mbox{s.t. }}

\newcommand*\xor{\mathbin{\oplus}}
\def\kui{\textcolor{black}}

\def\major{\textcolor{black}}

\usepackage{listings,xcolor} 
\usepackage{tcolorbox}

\newcommand{\lstfont}[1]{\color{#1}\ttfamily}

\usepackage{xspace}
\usepackage[colorlinks,linkcolor=red,anchorcolor=blue,citecolor=purple,CJKbookmarks=True]{hyperref}

\usepackage{marvosym}

\begin{document}

\title{Structured Binary Neural Networks for Image Recognition
}

\author{Bohan Zhuang,
        Chunhua Shen,
        Mingkui Tan,
        Peng Chen,
        Lingqiao Liu,
        and Ian Reid% <-this % stops a space}
}
\authorrunning{B. Zhuang, C. Shen, M. Tan, P. Chen,  L. Liu, I. Reid} % if too long for running head

\institute{
Bohan Zhuang, Peng Chen\at
              Faculty of Information Technology, Monash University \\
              \email{bohan.zhuang@monash.edu, blueardour@gmail.com}
           \and
           Chunhua Shen({\color{blue}{\Letter}}) \at
           %College of Computer Science and Technology,
           Zhejiang University \\
           \email{chunhuashen@zju.edu.cn} 
           \and
           Mingkui Tan \at School of Software Engineering, South China University of Technology
               \\
              \email{mingkuitan@scut.edu.cn}   \and
           Lingqiao Liu, Ian Reid \at
              School of Computer Science, The University of Adelaide\\
              \email{\{lingqiao.liu, ian.reid\}@adelaide.edu.au}
\\[0.2cm]
\it Accepted to: Int. J. Computer Vision, June 2022. 
}

%\date{Received: date / Accepted: date}
\date{\today}

\maketitle

\maketitle

\begin{abstract}
In this paper, we propose to train binarized convolutional neural networks (CNNs) that 
are of significant importance for deploying deep learning to 
mobile devices with limited  power capacity and computing resources. Previous works on quantizing CNNs often seek to approximate the floating-point information of weights and/or activations using a set of discrete values. Such methods, 
termed value approximation here, typically are built on the same network architecture 
of 
the full-precision
counterpart. 
Instead,  we take a new
``structured approximation'' view for network quantization --- it is possible
and valuable to exploit flexible architecture transformation when learning low-bit networks, which can achieve even better performance than the original networks in some cases. In particular, we propose a ``group decomposition'' strategy,
termed
GroupNet, which divides a network into desired groups. Interestingly, 
with
our GroupNet strategy, each full-precision group can be effectively reconstructed by aggregating a set of homogeneous binary branches. We also propose to learn effective connections among groups to improve the representation capability. 
\major{To improve the model capacity, we propose to dynamically execute sparse binary branches conditioned on input features while preserving the computational cost.}
More importantly, the proposed GroupNet shows strong flexibility %to
for a few vision
tasks. For instance, we extend the GroupNet for accurate semantic segmentation by embedding 
the 
rich context into the binary structure. The proposed GroupNet also shows strong 
performance on object detection. 
Experiments on image classification, semantic segmentation, and object detection tasks demonstrate the superior performance of the proposed methods over various quantized networks in the literature. 
Moreover, the speedup and runtime memory cost evaluation comparing with related quantization strategies is analyzed on GPU platforms, which serves as a strong benchmark for further research.

\keywords{Binary neural networks \and quantization \and image classification \and semantic segmentation \and object detection}
% \PACS{PACS code1 \and PACS code2 \and more}
% \subclass{MSC code1 \and MSC code2 \and more}
\end{abstract}

\section{Introduction}

Deep convolutional neural networks have achieved significant breakthroughs in many machine learning tasks, such as image classification~\cite{
he2016deep}, object segmentation~\cite{long2015fully, chen2018encoder} and object detection
\cite{tian2019fcos}.
However, deep models often require billions of FLOPs for inference, which makes them infeasible for many real-time applications especially on resource constrained mobile platforms. To solve this, 
existing works focus on network pruning~\cite{zhuang2018discrimination, he2017channel, lin2019towards}, low-bit quantization~\cite{zhuang2018towards, jacob2017quantization} and/or efficient architecture design \cite{chollet2017xception, howard2017mobilenets}. Among them, %the
quantization approaches seek to represent the weights and/or activations with low bitwidth fixed-point integers. 
Thus, 
the dot product can be computed by several XNOR-popcount bitwise operations. Binarization \cite{hubara2016binarized, rastegari2016xnor}, which is an extreme quantization approach, seeks to represent the weights and activations by a single bit (\textit{e.g.}, $+1$ or $-1$). Binarization has gained great attention recently since the XNOR of two bits 
can be executed extremely fast
\cite{ehliar2014area, govindu2004analysis}. In this paper, we aim to design highly accurate binary neural networks (BNNs) from a new quantization perspective.

Existing fixed-point quantization methods, including binarization, seek to quantize weights and/or activations by preserving most of the representational ability of the original network. In this sense,
these methods are based on the idea of \emph{value approximation},
which can be mainly divided into two categories. The first category
design
effective optimization algorithms to find better local minima for quantized weights.  These works either introduce knowledge distillation~\cite{zhuang2018towards, polino2018model, mishra2018apprentice} or use loss-aware objectives~\cite{hou2018loss, hou2017loss}. The second category approaches focus on improving the quantizer~\cite{zhou2016dorefa, Cai_2017_CVPR, zhang2018lq}, by learning suitable mappings between discrete values and their floating-point counterparts.

Note that, 
these
value approximation based approaches may have a natural limitation as 
it is merely a 
sub-optimal 
approximation to the original network.  Moreover, designing a good quantization function is highly non-trivial especially for BNNs, since the quantization process essentially is non-differentiable and the gradients can only be roughly approximated. Last, these methods often lack of adaptive ability for 
other
vision tasks beyond image classification.
Often these methods may work well on image classification tasks, but may not achieve promising quantization performance on segmentation and detection tasks.

In this paper, we investigate the task of quantization from a new perspective 
of 
\emph{structured approximation}. We
observe 
that, instead of directly approximating the original network, it is possible, and 
valuable to 
learn an ensemble of a set of binary bases
that can match the representational capability of the floating-point model. In particular, we propose a ``group decomposition" strategy 
termed
GroupNet, which 
partitions a full-precision model into groups. 
For the sake of exposition, 
we
use 
terminology for network architectures as comprising layers, blocks and groups.  A layer is a standard single parameterized layer in a network such as a dense or convolutional layer, except with binary weights and activations. A block is a collection of layers in which the output of end layer is connected to the input of the next block (\textit{e.g.}, a residual block). A group is a collection of blocks. In particular, one of the key points is, based on the proposed GroupNet strategy, we are able to use a set of binary bases to well approximate the floating-point model. 
This
design
shows
three benefits.

First, GroupNet enables more flexible trade-off between computational complexity and accuracy. Specifically, GroupNet enables fine-grained quantization levels that can be any positive integers (except 1) while fixed-point methods \cite{zhuang2018towards, zhou2016dorefa} require the quantization levels to be exponential power of 2. As a result, GroupNet can achieve the fine-grained bit-width by directly controlling the number of bases, which 
works better 
for balancing the efficiency and accuracy of the overall network.

Second, skip connections have been shown to be important in increasing representational power and improving gradient backpropagation as demonstrated in BNNs literature \cite{Liu_2018_ECCV}. In this sense, our group-wise design enjoys $K$ times more skip connections than value approximation, where $K$ is the number of bases.

Third, the higher-level structural information can be better utilized than the value approximation approaches.  
In practice, while the value approximation based approaches show promising performance on image classification tasks, they often perform poorly on more challenging tasks such as semantic segmentation and object detection.
Relying on the proposed group decomposition strategy, we are able to exploit task-specific information or structures and further design flexible binary structures according to specific tasks to compensate the quantization loss for general tasks.

Specifically,
for
semantic segmentation,
we are motivated by Atrous Spatial Pyramid Pooling (ASPP)~\cite{chen2017rethinking, chen2018encoder}, which is built on top of extracted features of the backbone network. To capture the multi-scale context, we propose to directly apply different atrous rates on parallel binary bases in the backbone network, which is equivalent to absorbing ASPP into the feature extraction stage. As will be shown, our strategy significantly boosts the performance on semantic segmentation without increasing much computational complexity of the binary convolutions. Moreover, we further extend the proposed approach to building quantized networks for object detection. Building low-precision networks for object detection is more challenging since detection needs the network to output richer information, 
locations and categories of bounding boxes. 
Several works in literature
address 
quantized object detection \cite{jacob2017quantization, li2019fully, wei2018quantization}. 
There is still a considerable performance gap between quantized object detectors and
their full-precision counterparts. 
To tackle this problem, we apply our GroupNet and propose a new design modification to better accommodate the quantized object detector and achieve improved detection accuracy.
Last, it is worth mentioning that our structured approximation strategy and the value approximation strategy are complementary rather than contradictory. In fact, both are important and should be exploited to obtain highly accurate BNNs.

In this paper, we propose
to redesign binary network architectures from the quantization view. We highlight that while most existing quantization works focus on directly quantizing the full-precision architecture, 
we
begin to explore alternative architectures that shall be better suited for  dealing with binary weights and activations.
In particular, apart from decomposing each group into several binary bases, we also propose to learn the connections between each group by introducing a soft fusion gate. To further increase model capacity while preserving the computational cost, we propose to tailor conditional computing to binary networks by learning to select informative bases for each group conditioned on input features.

Our main contributions are summarized as follows.
	\begin{itemize}
    \item{We propose to design accurate BNNs structures from the \emph{structured approximation} perspective. Specifically, we divide the network into groups and approximate each group using a set of binary bases. We also propose to automatically learn the decomposition by introducing soft connections.}
    
    \item{\major{We explore ideas from conditional computing to learn 
    adaptive, conditional 
    binary bases, which are dynamically selected during inference conditioned on the input features. This strategy significantly increases the model capacity of GroupNet without increasing computational complexity during inference.}}

    \item{
    The proposed GroupNet has strong flexibility and can be easily extended to tasks other than image classification. For instance, 
    we propose Binary Parallel Atrous Convolution (BPAC), which embeds rich multi-scale context into BNNs for 
    semantic segmentation. 
    GroupNet with BPAC significantly improves the performance while maintaining the complexity compared to employ GroupNet only.}
    \item{To
    our knowledge, we 
    are 
    among the pioneering approaches to apply binary neural networks to general semantic segmentation and object detection tasks.}
    \item{We develop 
    implementations to evaluate the execution speed and runtime memory cost of GroupNet and make comparison with other bit configurations on various platforms.}
    \item{We evaluate our models on ImageNet, PASCAL VOC and COCO datasets based on various architectures. 
    Experiments show that the proposed Gr\-ou\-p\-Net a\-ch\-ie\-ve\-s the state-of-the-art trade-off between accuracy and computational complexity.} 
	\end{itemize}

This paper extends our preliminary results 
in 
\cite{zhuang2019structured}
in several aspects. 
1) In the conference version, we propose to employ the soft routing mechanism to learn group-wise connections. Nevertheless, all the branches still need to be executed at test time. In this paper, we further explore more advanced structured approximation for BNNs. As opposed to static ones, we adapt the structures to the input during inference, with enlarged parameter space and improved model capacity. In particular, we employ conditional computation by optimizing data-dependent binary gates to decide the execution of branches in each group.
2) We develop the acceleration code on resource constrained platforms and evaluate speedup and runtime memory cost comparing with various quantization methods.
3) We make more analysis on differences and advantages of GroupNet over other related quantization strategies.
4) In addition to image classification and semantic segmentation, we further extend GroupNet to object detection. In particular, we propose several modifications to quantized object detection and our GroupNet outperforms the comparison methods.
5) For image classification, we conduct more ablation studies and experiments on more architectures and provide detailed analysis. 
6) For semantic segmentation, we develop models based on DeepLabv3 and provide useful instructions.

\section{Related Work}
\textbf{Network quantization.} The recent increasing demand for implementing fixed-point deep neural networks on embedded devices motivates the study of low-bit network quantization. 
Quantization based methods represent the network weights and/or activations with very low precision, thus yielding highly compact DNN models compared to their floating-point counterparts.
BNNs \cite{hubara2016binarized, rastegari2016xnor} propose to constrain both weights and activations to binary values (i.e., $+1$ and $-1$), where the multiplication-accumulations can be replaced by purely $\rm xnor(\cdot)$ and $\rm popcount(\cdot)$ operations, which are in general much faster. 
However, BNNs still suffer from significant accuracy 
decrease 
compared with the full precision counterparts. 
To narrow this accuracy gap, 
ternary networks \cite{li2016ternary,zhu2016trained} and even higher bit fixed-point quantization \cite{zhou2016dorefa, zhou2017incremental} methods are proposed.

In general, quantization approaches target at tackling two main problems. On one hand, some works target at designing a more accurate quantizer to minimize information loss. For the uniform quantizer, works in \cite{choi2018pact,jung2019learning} explicitly parameterize and optimize the upper and/or lower bound of the activation and weights. To reduce the quantization error, non-uniform approaches \cite{park2017weighted, zhang2018lq} are proposed to better approximate the data distribution. 
In particular, LQ-Net \cite{zhang2018lq} proposes to jointly optimize the quantizer and the network parameters. 
On the other hand, because of the non-differentiable quantizer, some literature focuses on relaxing the discrete optimization problem. A typical approach is to train with regularization \cite{ding2019regularizing, bai2019proxquant}, where the optimization problem becomes continuous while gradually adjusting the data distribution towards quantization level. Moreover, Hou \etal
\cite{hou2017loss, hou2018loss} propose the loss-aware quantization by directly optimizing the discrete objective function.

To well balance accuracy and complexity, several works~\cite{guo2017network, li2017performance, lin2017towards, fromm2018heterogeneous, tang2017train} propose to employ a linear combination of binary tensors to approximate the filters and/or activations while still possessing the advantage of binary operations. In particular, Guo~\etal\cite{guo2017network} recursively perform residual quantization on pretrained full-precision weights and do convolution on each binary weight base. Similarly, Li~\etal\cite{li2017performance} propose to expand the input feature maps into binary bases in the same manner.
%And
%
Lin~\etal\cite{lin2017towards} further expand both weights and activations with a simple linear approach.
Unlike the previous local tensor approximation approaches, we propose to design BNNs from a structured approximation perspective and show strong generalization on a few mainstream computer vision tasks.

There also have been several works that employ neural architecture search (NAS) for BNNs \cite{bulat2020bats, zhu2020nasb, chen2020binarized, ding2020bnas} to explore a high-performance binary neural architecture. Note that our GroupNet is orthogonal to these approaches, where we explore how to ensemble the binary bases effectively to approximate the original full-precision network. Based on an advanced binary base architecture, our GroupNet can achieve better performance.

\noindent\textbf{Hardware Implementation.}
In addition to the quantization algorithms design, the implementation frameworks and acceleration libraries \cite{ignatov2018ai,chen2018tvm,umuroglu2017finn,jacob2017quantization, yang2017bmxnet} are indispensable to expedite the quantization technique to be deployed on energy-efficient edge devices. For example, TBN \cite{Wan_2018_ECCV} focuses on the implementation of ternary activation and binary weight networks. daBNN \cite{zhang2019dabnn} targets at the inference optimization of BNNs on ARM CPU devices.
GXNOR-Net \cite{deng2018gxnor} treats TNNs as 
sparse BNNs and propose an acceleration solution on dedicated hardware platforms. In this paper, we develop the acceleration code for BNNs, GroupNet and fixed-point quantization on GPU platforms. We also compare the accuracy and efficiency trade-offs between them.

\noindent\textbf{Dynamic networks.}
Dynamic networks, as opposed to static ones, can adapt their 
structures or parameters to the input during inference, and therefore enjoy desired trade-off between accuracy and efficiency for dealing with varying computational budgets on the fly. In particular, closely related to our paper, mixture of experts (MoE) \cite{jacobs1991adaptive, jordan1994hierarchical} build multiple network branches as experts in parallel. For example, HydraNet \cite{mullapudi2018hydranets} replaces the convolutional blocks in the last stage of a CNN by multiple branches, and selectively execute these branches at test time.
The recent Switch Transformer \cite{fedus2021switch} proposes to dynamically activate a FFN layer from experts, leading to trillion parameter models with constant computational cost. In this paper, we tailor conditional computing to GroupNet by learning to select data-specific bases for each group. 
Related to our GroupNet, High-capacity Expert BNNs (HCE) \cite{bulat2021high} proposes to automatically search for optimal binary network architectures, equipped with multiple experts to increase model capacity. During inference, HCE only selects one expert, while our GroupNet enables Top-$N$ selection for larger model capacity. Moreover, our GroupNet focuses on structured experts, while the scope of HCE is on designing layer-wise experts for binary convolutions.

\noindent\textbf{Semantic segmentation.}
Deep learning based semantic segmentation 
is popularized by the Fully Convolutional Networks (FCNs)~\cite{long2015fully}.  
Recent prominent directions have emerged: using the encoder-encoder structure~\cite{zhang2018exfuse, chen2018encoder}; relying on dilated convolutions to keep the reception fields without downsampling the spatial resolution too much~\cite{chen2017rethinking, mehta2018espnet}; employing multi-scale feature fusion~\cite{lin2017refinenet, chen2018deeplab}; 
employment of probabilistic
graphical models~\cite{lin2016efficient,chandra2016fast}. 

However, these approaches typically focus on designing complex modules for improving accuracy while sacrificing the inference efficiency to some extent.
To make semantic segmentation applicable, 
several
methods have been 
proposed to design real-time semantic segmentation models. 
Recently, works of \cite{liu2019auto, nekrasov2019fast} apply neural architecture search for exploring more accurate models with less Multi-Adds operations. Yu \etal \cite{yu2018bisenet} propose BiSeNet, where a spatial path extracts high resolution features and a context path obtains sufficient receptive fields to achieve high speed and accuracy. ESPNet~\cite{mehta2018espnet, mehta2018espnetv2} design efficient spatial pyramid for real-time semantic segmentation under resource constraints.
In contrast, we instead propose to accelerate semantic segmentation frameworks from the quantization perspective, which is parallel to the above approaches.
Given a pretrained full-precision model, we can replace multiplication-accumulations by the XNOR-popcount operations, which would bring great benefits for embedded platforms. We may be the first to apply binary neural networks on semantic segmentation and achieve promising results.

\noindent\textbf{Object detection.}
Object detection has shown great success with deep neural networks. As one of the dominant detection framework, two-stage detection methods~\cite{girshick2015fast, girshick2014rich, ren2015faster} first generate region proposals and then refine them by subsequent networks. The popular method Faster-RCNN \cite{ren2015faster} first proposes an end-to-end detection framework by introducing a region proposal network (RPN). 
Another main category is the one-stage methods which are represented by YOLO~\cite{redmon2016you, redmon2017yolo9000, redmon2018yolov3}, SSD \cite{liu2016ssd} and FCOS \cite{tian2019fcos}. The objective is to improve the detection efficiency by directly classifying and regressing the pre-defined anchors without the proposal generation step. RetinaNet~\cite{lin2017focal} proposes a new focal loss to tackle the extreme foreground-background class imbalance encountered during training in one-stage detectors. Moreover, Tian \etal \cite{tian2019fcos} propose a simple fully convolutional anchor-free one-stage detector that achieves 
\textit{on par}
performance with the anchor-based one-stage detectors.

In addition, designing light-weight detection frameworks is crucial since mobile applications usually require real-time, low-power and fully embeddable. For example, the work of \cite{chen2017learning} and \cite{wei2018quantization} propose to train a tiny model by distilling knowledge from a deeper teacher network. MNasNet~\cite{tan2018mnasnet} proposes to automatically search for mobile CNNs which achieve 
improved 
mAP quality than MobileNets for 
object detection.
In this paper, we explore to design efficient detectors from the quantization view.
Note that, 
we are probably the first to train a binary object detection model.

\section{Proposed Method}
\noindent {Most previous methods in the literature have focused on value approximation by designing accurate binarization functions for weights and activations (e.g., multiple binarizations~\cite{lin2017towards, tang2017train, guo2017network, li2017performance, fromm2018heterogeneous}).} {In this paper, we seek to binarize both weights and activations of CNNs from a ``group-wise approximation" view.} 
In the following, we first define the
research problem 
and present some basic knowledge about binarization in Section~\ref{sec:function}.
Then, we explain our binary architecture design strategy in Section~\ref{sec:decomposition} and introduce how to learn architectures automatically and dynamically in Section \ref{sec:learn}.
In Section \ref{sec:method_complexity}, we further provide the complexity analysis. 
Finally, in Section~\ref{sec:segmentation} and Section~\ref{sec:detection}, we describe how to use task-specific attributes to generalize our approach to semantic segmentation and object detection, respectively.

\subsection{Problem definition}\label{sec:function}
\noindent For a convolutional layer, we define the input feature ${\bf{x}} \in {\mathbb{R}^{{c_{in}} \times {w_{in}} \times {h_{in}}}}$, weight tensor ${\bf{w}} \in {\mathbb{R}^{{c_{in}} \times {c_{out}}  \times w \times h}}$ and the output ${\bf{y}} \in {\mathbb{R}^{{c_{out}} \times {w_{out}} \times {h_{out}}}}$, respectively.

\noindent\textbf{Binarization of weights}: Following ~\cite{rastegari2016xnor}, we approximate the floating-point weight ${\bf{w}}$ by a binary weight filter ${{\bf{b}}^w}$ and a scaling factor $\alpha  \in {\mathbb{R}^ + }$ such that ${\bf{w}} \approx \alpha {{\bf{b}}^w}$, where ${{\bf{b}}^w}$ is the sign of ${\bf{w}}$ and $\alpha$ calculates the mean of absolute values of ${\bf{w}}$. In general, $\rm sign(\cdot)$ is non-differentiable and so we adopt the straight-through estimator~\cite{bengio2013estimating} (STE) to approximate the gradient calculation. 
Formally, the forward and backward processes can be 
written 
as follows: 
\begin{equation}  \label{eq:binary_weights}
\begin{split}
&\mbox{Forward}:{{\bf{b}}^w} = {\rm{sign}}({\bf{w}}),\\
&\mbox{Backward}:\frac{{\partial \ell }}{{\partial {\bf{w}}}} = \frac{{\partial \ell }}{{\partial {{\bf{b}}^w}}} \cdot \frac{{\partial {{\bf{b}}^w}}}{{\partial {\bf{w}}}} \approx \frac{{\partial \ell }}{{\partial {{\bf{b}}^w}}},\\
\end{split}
\end{equation}
where $\ell$ is the loss.

\noindent\textbf{Binarization of activations}: 
For activation binarization, we utilize the piecewise polynomial function to approximate the sign function as in~\cite{Liu_2018_ECCV}. The forward and backward can be written as:
\begin{equation} \label{eq:binary_activations}
\begin{array}{l}
\mbox{Forward}: {b^a} = {\rm{sign}}(x),\\
\mbox{Backward}:\frac{{\partial \ell }}{{\partial x}} = \frac{{\partial \ell }}{{\partial {b^a}}} \cdot \frac{{\partial {b^a}}}{{\partial x}},\\
\mbox{where} \quad \frac{{\partial {b^a}}}{{\partial x}} = \left\{ \begin{array}{l}
2 + 2x: - 1 \le x < 0\\
2 - 2x:0 \le x < 1\\
0:\rm{otherwise}
\end{array} \right..
\end{array}
\end{equation}

\subsection{Binary Network  Decomposition} \label{sec:decomposition}

\noindent \kui{In this paper, we want to  design a new structural representation of a network for quantization. } {First of all, note that a float number in computer is represented by a fixed-number of binary digits. Motivated by this, rather than directly doing the quantization via ``value decomposition", we propose to decompose a network  into binary structures while preserving its representability.} 

Specifically, given a floating-point residual network $\Phi(\cdot)$ with $N$ blocks,  we decompose $\Phi(\cdot)$ into $P$ binary fragments $[{{\cal F}_1}, ..., {{\cal F}_P}]$, 
where ${{\cal F}_i}(\cdot)$ can be any binary structure. Note that each ${{\cal F}_i}(\cdot)$ can be different. \kui{A natural question arises: can we find some simple  methods to decompose the network with binary structures so that the representability can be exactly preserved? To answer this question, we here explore two kinds of architectures}  for ${{\cal F}}(\cdot)$, namely layer-wise decomposition and group-wise decomposition in Section~\ref{sec:layerwise} and Section~\ref{sec:groupwise}, respectively. 
% After that,
Then we %will
present a novel strategy for automatic decomposition in Section~\ref{sec:learn}.

\begin{figure*}
	\centering
	\resizebox{1.0\linewidth}{!}
	{
		\begin{tabular}{c}
			\includegraphics{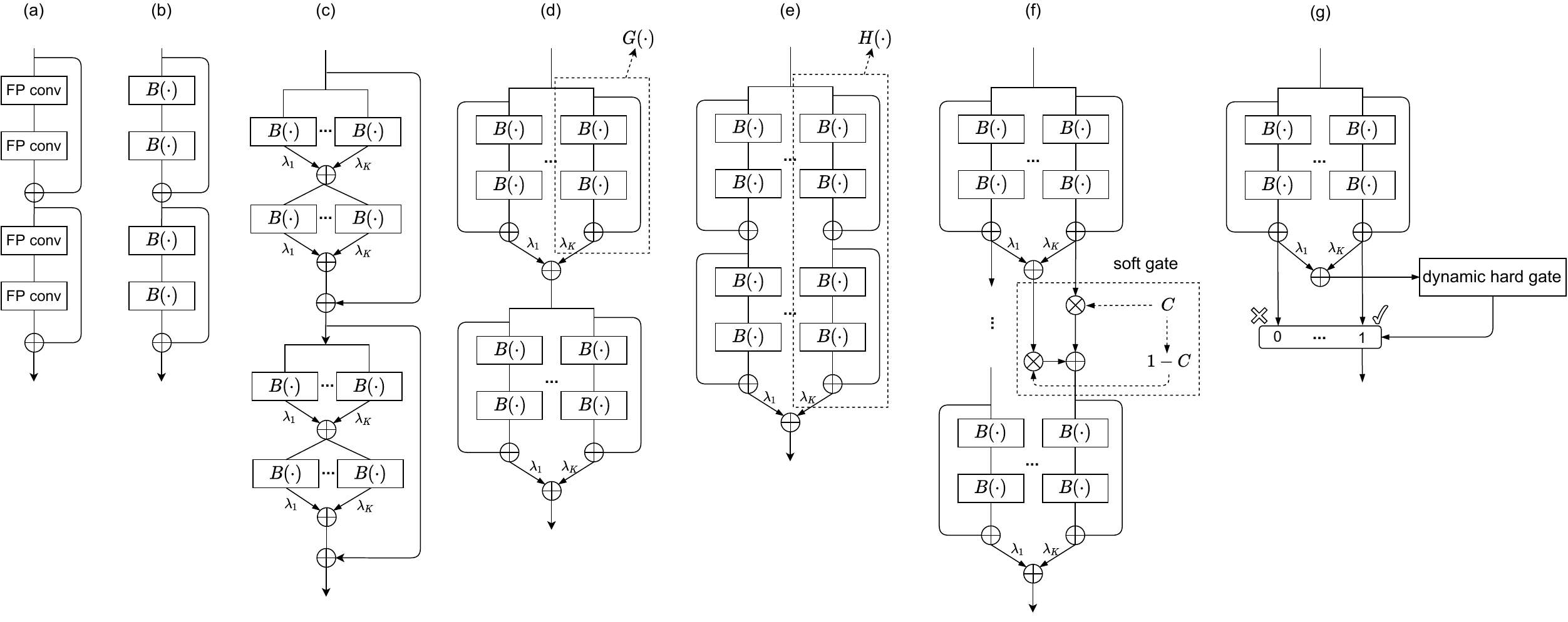}
		\end{tabular}
	}
	\caption{\major{Illustration of the proposed method.} 
We take one residual block with two convolutional layers for illustration. For convenience, we omit batch normalization and nonlinearities. (a): The floating-point residual block. (b): Direct binarization of a full-precision block. (c): Layer-wise binary decomposition in Eq.~(\ref{eq:1}), where we use a set of binary convolutional layers $B(\cdot)$ to approximate a floating-point convolutional layer. (d): Basic group-wise binary decomposition in Eq.~(\ref{eq:2}), where we approximate a whole block with a linear combination of binary blocks $G(\cdot)$. (e): We approximate a whole group with homogeneous binary bases $H(\cdot)$, where each group consists of several blocks. This corresponds to Eq.~(\ref{eq:3}). (f): Illustration of the soft connection between two neighbouring blocks. (g): Hard gating mechanism for dynamically selecting branches.}
%    \vspace{-1.0em}
	\label{fig:main}
\end{figure*}

\subsubsection{Layer-wise binary decomposition} \label{sec:layerwise}

The key challenge of binary decomposition is how to reconstruct or approximate the floating-point structure. The simplest way is to approximate in a layer-wise manner. Let $B(\cdot)$ be a binary convolutional layer and ${\bf{b}}_i^w$ be the binarized weights for the $i$-th base. In Figure~\ref{fig:main}(c), we illustrate the layer-wise feature reconstruction for a single block.
Specifically, for each layer, we aim to fit the full-precision structure using a set of binarized homogeneous branches ${\cal F(\cdot)}$ given a floating-point input tensor $\bf{x}$:
\begin{equation}\label{eq:1}
{{\cal F}}({\bf{x}}) = \frac{1}{K}\sum\limits_{i = 1}^K {B_i}({\bf{x}})  = \frac{1}{K}\sum\limits_{i = 1}^K ({\bf{b}}_i^w \oplus {\rm{sign}}({\bf{x}})),
\end{equation} 
where $\oplus$ is bitwise operations $\rm xnor(\cdot)$ and $\rm popcount(\cdot)$, $K$ is the number of branches. 
During the training, the structure is fixed and each binary convolutional kernel ${\bf{b}}_i^w$ is directly updated with end-to-end optimization.  The scale scalar can be absorbed into batch normalization when doing inference. 
\kui{Note that all ${{B_i}}$'s in Eq.~(\ref{eq:1}) have the same topology as the original floating-point counterpart.} Each binary branch gives a rough approximation and all the approximations are aggregated to achieve more accurate reconstruction to the original full precision convolutional layer. Note that when $K=1$, it corresponds to directly binarize the floating-point convolutional layer (Figure~\ref{fig:main} (b)). However, with more branches (a larger $K$), we are expected to achieve more accurate approximation with more complex transformations. 

Different from~\cite{zhuang2019structured}, we remove the floating-point scales for two reasons.
First, the scales can be absorbed into batch normalization layers. Second, we empirically observe that the learnt scales for different branches differ a little and removing them does not influence the performance.

During the inference, the homogeneous $K$ bases can be parallelizable and thus the structure is hardware friendly. This will bring significant gain in  speed-up of the inference. 
Specifically, the bitwise XNOR operation and bit-counting can be performed in a parallel of 64 by the current generation of CPUs~\cite{rastegari2016xnor, Liu_2018_ECCV}. 
We 
only 
need to calculate $K$ binary convolutions and $K$ full-precision additions. As a result, the speed-up ratio $\sigma$ for a convolutional layer can be calculated as:
\begin{equation}
\sigma  =  \frac{{ 64 
{c_{{in}}} \cdot  w  \cdot  h  \cdot {w_{in}} \cdot {h_{in}}}}
{K({{c_{{in}}} \cdot w \cdot h 
\cdot
{w_{in}} \cdot  {h_{in}} + 64{w_{out}}
\cdot 
{h_{out}}})} .
\end{equation}
We take one layer in ResNet for example. If we set ${c_{in}} = 256$, $w \times h = 3 \times 3$, ${w_{in}}= {h_{in}}= {w_{out}}= {h_{out}} = 28$, $K=5$, then it can reach 12.45$\times$ speedup. In practice, the actual speedup ratio is also influenced by the process of memory read and thread communication. We extensively evaluate the on-device speedup in Section \ref{sec:speedup}.

\subsubsection{Group-wise binary decomposition}\label{sec:groupwise}

In {the layer-wise approach, we approximate each layer with multiple branches of binary layers. Note each branch will introduce a certain amount of error and  the error may accumulate due to the aggregation of multi-branches. As a result, this strategy  may incur severe quantization errors and bring large deviation for gradients during back-propagation.} 
To alleviate the above issue, we further propose a more flexible decomposition strategy called group-wise binary decomposition, to preserve more structural information during approximation.

To explore the group-structure decomposition, we first consider {a simple} case where each group consists of only one block. {Then,} the layer-wise approximation strategy can be easily extended to the group-wise case. As shown in Figure~\ref{fig:main} (d), similar to the layer-wise case, each floating-point group is decomposed into multiple binary groups. However, each group ${G_i(\cdot)}$ is a binary block which consists of several binary convolutions and fixed-point operations (i.e., AddTensor). For example, we can set ${G_i(\cdot)}$ as the basic residual block~\cite{he2016deep} which is shown in Figure~\ref{fig:main} (a).
Considering the residual architecture, we can decompose ${\cal F}({\bf{x}}) $  by extending Eq.~(\ref{eq:1}) as:
\setlength{\abovedisplayskip}{4pt} 
\setlength{\belowdisplayskip}{4pt}
\begin{equation} \label{eq:2}
{\cal F}({\bf{x}}) = \frac{1}{K}\sum\limits_{i = 1}^K {G _i}({\bf{x}}).
\end{equation}
In Eq.~(\ref{eq:2}), we use a linear combination of homogeneous binary bases to approximate one group, where each base ${G _i}$ is a binarized block. 
Thus,
we effectively keep the original residual structure in each base to preserve the network capacity. 
As shown in Section~\ref{sec:group_space}, the group-wise decomposition strategy performs much better than the simple layer-wise approximation.

Furthermore, the group-wise approximation is flexible. We now analyze the case where each group may contain different number of blocks. Suppose we partition the network into $P$ groups and it follows a simple rule that each group must include one or multiple complete residual building blocks.
For the $p$-th group, we consider the blocks set $T \in \{ {T_{p - 1}} + 1,...,{T_p}\}$, where the index ${T_{p - 1}}=0$ if $p=1$.  We can extend Eq.~(\ref{eq:2}) into multiple blocks format:
\begin{equation}\label{eq:3}
\begin{array}{l}
{\cal F}({{\bf{x}}_{{T_{p - 1}} + 1}}) = \frac{1}{K}\sum\limits_{i = 1}^K {H_i}({\bf{x}}),\\
 = \frac{1}{K}\sum\limits_{i = 1}^K {G_i^{{T_p}}(G_i^{{T_p} - 1}(...(G_i^{{T_{p - 1}} + 1}({{\bf{x}}_{{T_{p - 1}} + 1}}))...)} ),
\end{array}
\end{equation}
where $H(\cdot)$ is a cascade of consequent blocks which is shown in Figure~\ref{fig:main} (e).
Based on ${\cal{F(\cdot)}}$, we can efficiently construct a network by stacking these groups and each group may consist of one or multiple blocks. Different from Eq.~(\ref{eq:2}), we further expose a new dimension on each base, which is the number of blocks. This greatly increases the structure space and the flexibility of decomposition. 
We explore the effect of the structure space in Section \ref{sec:group_space}
and further describe how to learn the decomposition in Section~\ref{sec:learn}.

\subsection{Learning for decomposition} \label{sec:learn}
\noindent
There is a significant challenge involved in Eq.~(\ref{eq:3}). Note that the network has $N$ blocks and the possible number of connections is ${2^N}$. Clearly, it is not practical to enumerate all possible structures during the training. Here, we propose to solve this problem by  learning the structures for decomposition automatically.
\subsubsection{Learned soft gating} \label{sec:soft_gating}
We introduce in a fusion gate as the soft connection between blocks $G(\cdot)$. To this end, we first define the input of the $i$-th branch for the $n$-th block as:
\begin{equation} \label{eq:decompose}
\begin{aligned}
C_i^n &= \mathrm{sigmoid}({\theta ^n _i}),\\
{\bf{x}}_i^n&= {C ^n _i} \odot G_i^{n - 1}({\bf{x}}_i^{n - 1})  \\&+ (1 - {C^n_i}) \odot \sum\limits_{j = 1}^K G_j^{n - 1}({\bf{x}}_j^{n - 1}),
\end{aligned}
\end{equation}
where $\btheta  \in {\mathbb{R}^K}$ is a learnable parameter vector, $C_i^n$ is a gate scalar and $\odot$ is the Hadamard product. And we empirically observe that using a learnable scale $\theta$ that shares among branches does not influence the performance and it can be absorbed into batch normalization (BN) layers during inference.

Here, the branch input ${\bf{x}}_i^n$ is a weighted combination of two paths. The first path is the output of the corresponding $i$-th branch in the $(n-1)$-th block, which is a direct connection. The second path is the aggregation output of the $(n - 1)$-th block. The detailed structure is shown in Figure~\ref{fig:main} (f). In this way, we make more information flow into the branch and increase the gradient paths for improving the convergence of BNNs.

{\noindent\textbf{Remarks:} For the extreme case when { {$\sum\nolimits_{i = 1}^K {C_i^n} = 0$}}, Eq.~(\ref{eq:decompose}) will be reduced to Eq.~(\ref{eq:2}) which means we approximate the $(n-1)$-th and the $n$-th block independently.} When {{$\sum\nolimits_{i = 1}^K {{C_i^n}} = K$}},  Eq.~(\ref{eq:decompose})  is equivalent to Eq.~(\ref{eq:3})  and we set $H(\cdot)$ to be two consequent blocks and approximate the group as a whole.
Interestingly, when {{$\sum\nolimits_{n = 1}^N {\sum\nolimits_{i = 1}^K {C_i^n} } = NK$}}, it corresponds to set $H(\cdot)$ in Eq.~(\ref{eq:3}) to be a whole network and directly ensemble $K$ binary models.

Note that the group-wise decomposition has $K \times$ more skip connections than the layer-wise decomposition, implying the improved representational power and gradient backpropagation of the model.

\subsubsection{\major{Dynamic hard gating}}  \label{sec:hard_gating}

The soft gating regime in Section \ref{sec:soft_gating} adopts real-valued weights to dynamically rescale the representations obtained from different branches. In this way, all the branches still need to be executed, and the computation cannot be reduced at test time. To improve the inference efficiency, we propose to learn dynamic hard gates, allowing the models to adaptively allocate the computation dependent on the input features with only a fraction of activated branches. Such a design is shown in Figure~\ref{fig:main} (g).
Formally, the gating mechanism is applied to the output of the preceding group and can be formulated as 
\begin{equation} \label{eq:hard_gate}
g_i = \begin{cases}
1, & \text { if } \psi(\bx)_i \geq f(\psi(\bx), N) \\ 0, & \text { otherwise }\end{cases}
\end{equation}
where $g_i$ is the $i$-th element of
a $K$-dimensional binary vector $\bg$ which is used to select the $N$ activated branches, $\psi(\cdot)$ is a learned mapping function of the input $\bx$ and $f(\psi(\bx), N)$ returns the $N$-th largest element in $\psi(\bx)$.

The light-weight projection $\psi(\cdot)$ is defined as:
\begin{equation}
    \psi(\bx) = \mathrm{mean}(\bx)\boldsymbol{\nu},
\end{equation}
where $\mathrm{mean}(\cdot)$ is the channel-wise average pooling for feature summary and $\boldsymbol{\nu} \in \mathbb{R}^{c_{in} \times K}$ is a trainable linear transformation.

Nevertheless, during training, the selection process in Eq. (\ref{eq:hard_gate}) is non-differentiable. Moreover, it is crucial to update the parameters of the non-selective branches. To address this, we use $\text{Softmax}(\cdot)$ to approximate the gradients of $\bg$ for the back propagation:
\begin{equation}
  \frac{{\partial \bg}}{{\partial \bx}} \approx \frac{{\partial \text{Softmax}(\bx) }}{{\partial \bx}}.  
\end{equation}
In this way, we effectively address the gradient mismatch between the forward and backward, while allowing all experts are updated properly during training.

\subsection{Complexity and model size analysis} \label{sec:method_complexity}
\begin{table*}[!htb]
\centering
\caption{Computational complexity and static storage comparison of different quantization approaches. $F$: full-precision, $B$: binary, $Q_K$: $K$-bit quantization.}
\scalebox{0.9}
	{
		\begin{tabular}{r | c c c c c c c}
			\hline
			Model &Weights &Activations &Operations  &Model Size Saving &Computation Saving \\
			\hline
			Full-precision DNN &$F$  &$F$  &+, $-$, $\times$ &1  &1    \\\hline
			\cite{hubara2016binarized, rastegari2016xnor} &$B$ &$B$ &XNOR-popcount &$\sim 32 \times$ &$\sim 64 \times$   \\\hline
			\cite{courbariaux2015binaryconnect, hou2017loss} &$B$ &$F$  &+, $-$ &$\sim 32 \times$  &$\sim 2 \times $\\\hline
			\cite{zhu2016trained, zhou2017incremental} &$Q_K$  &$F$  &+, $-$, $\times$  &$\sim\frac{{32}}{K} \times$  &$< 2 \times$   \\\hline			
			\cite{zhuang2018towards, zhou2016dorefa} &$Q_K$ &$Q_K$ &+, $-$, $\times$  & $\sim\frac{{32}}{K} \times$ &$< \frac{{64}}{{{K^2}}} \times$  \\\hline
			\cite{lin2017towards, li2017performance} &$K \times B$  &$K \times B $ & +, $-$, XNOR-popcount &$\sim\frac{{32}}{K} \times$ & $ < \frac{{64}}{{{K^2}}} \times$  \\\hline
		    GroupNet &\multicolumn{2}{|c}{$K \times (B, B)$} &+,$ -$, XNOR-popcount  &$\sim\frac{{32}}{K} \times$ &$< \frac{{64}}{K} \times$ \\\hline 
			
	\end{tabular}}
%	\vspace{-1em}
	\label{tab:complexity_compare}
\end{table*}
A comprehensive comparison of various quantization approaches over complexity and model size is shown in Table~\ref{tab:complexity_compare}. For example, in some previous multiple binarization approaches \cite{liu2019circulant, lin2017towards}, each convolutional layer is approximated using $K$ weight bases and $K$ activation bases, which needs to calculate $K^2$ times binary convolution. In contrast, we just need to approximate several groups with $K$ structural bases. As reported in Section~\ref{sec:imagenet}, we save approximate $K$ times computational complexity while still achieving comparable Top-1 accuracy.
Since we use $K$ structural bases, the number of parameters increases by $K$ times in comparison to the full-precision counterpart. But we still save the model size by $\sim32/K$ times since the weights for convolutional layers are binary 
in our GroupNet, except for the 8-bit first and last layers, and the 8-bit $1 \times 1$ downsampling layers in some GroupNet variants.
For our approach, there exists element-wise operations between each group, so the computational complexity saving is slightly less than $\frac{{64}}{K} \times$.

\begin{figure}[htp]   
\centering
\resizebox{0.9\linewidth}{!}
{
	\begin{tabular}{c}
		\includegraphics{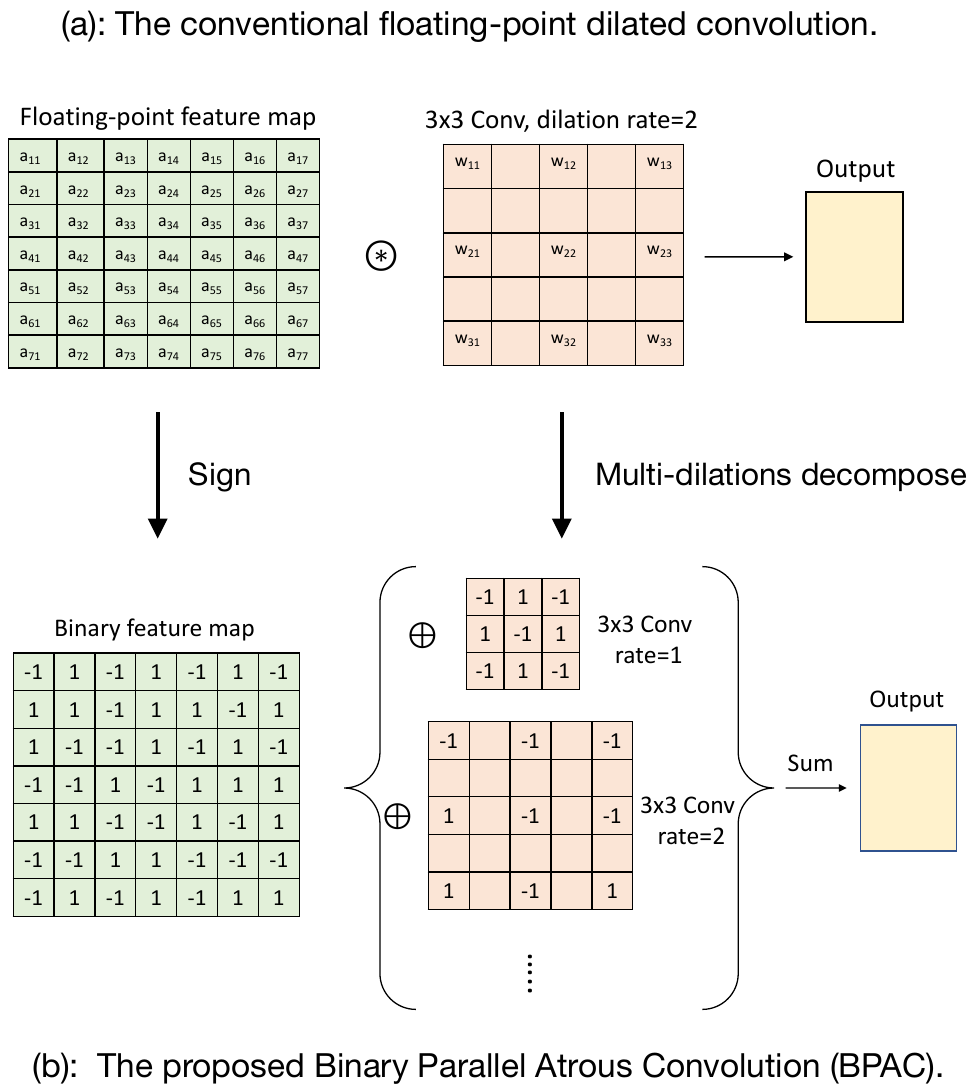}
	\end{tabular}
}
	
	\caption{The comparison between conventional dilated convolution and BPAC. For expression convenience, the group only has one convolutional layer. $\circledast$ is the convolution operation and $\oplus$ indicates the XNOR-popcount operations. (a): The original floating-point dilated convolution. (b): We decompose the floating-point atrous convolution into a combination of binary bases, where each base uses a different dilated rate. We sum the output features of each binary branch as the final representation.}

	\label{fig:segmentation}
\end{figure}
%\vspace{-1mm}
\subsection{Extension to semantic segmentation} \label{sec:segmentation}

\noindent The key message conveyed in the proposed method is that,
although each binary branch has a limited modeling capability, aggregating them together leads to a powerful model. In this section, we show that this principle can be applied to tasks other than image classification. In particular, we consider semantic segmentation which can be deemed as a dense pixel-wise classification problem. In the state-of-the-art semantic segmentation network, the atrous convolutional layer~\cite{chen2017rethinking} is an important building block, which performs convolution with a certain dilation rate. To directly apply the proposed method to such a layer, one can construct multiple binary atrous convolutional branches with the same structure and aggregate results from them.  However, we choose not to do this but propose an alternative strategy: we use different dilation rates for each branch. In this way, the model can leverage multiscale information as \textit{a by-product of the network branch decomposition.} It should be noted that this scheme does not incur any additional model parameters and computational complexity compared with the naive binary branch decomposition. The idea is illustrated in Figure~\ref{fig:segmentation} and we 
term 
this strategy Binary Parallel Atrous Convolution (\textbf{BPAC}).

In this work, we use the same ResNet backbone in~\cite{chen2017rethinking, chen2018encoder} with \emph{output\ stride=8}, where the last two stages employ atrous convolution. In BPAC, we keep $rates = \{2,...,K+1\}$ and $rates = \{6,...,K+5\}$ for $K$ bases in the last two blocks, respectively.
Intriguingly, as will be shown in Section~\ref{exp:pascal}, our strategy brings so much benefit that using five binary bases with BPAC
achieves similar performance as the original full precision network despite the fact that it saves considerable computational cost. 

\subsection{Extension to object detection} \label{sec:detection}

\begin{figure*}[tb]
	\centering
	\resizebox{.8471\linewidth}{!}
	{
		\begin{tabular}{c}
			\includegraphics{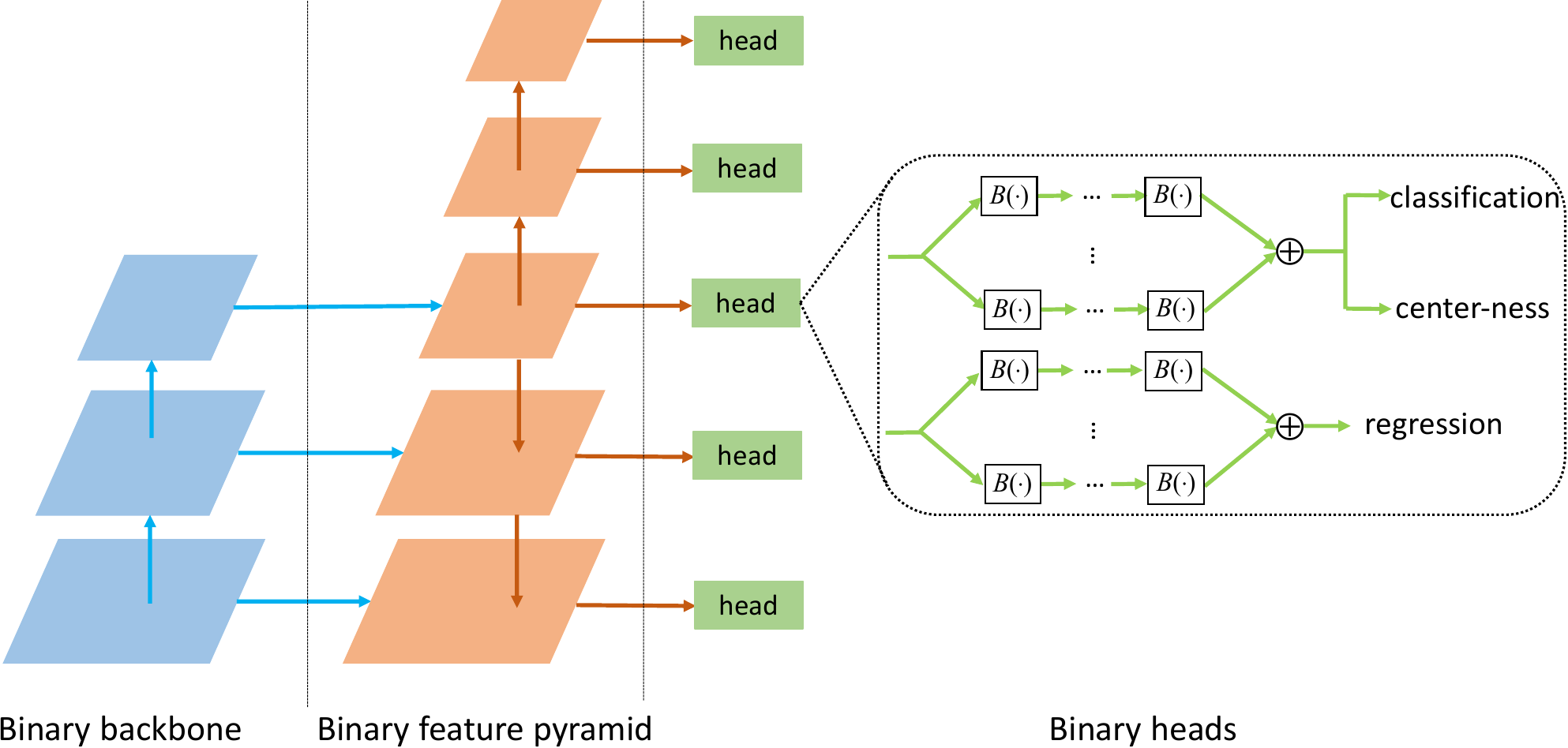}
		\end{tabular}
	}
	\caption{Illustration of the proposed binary detection framework. We partition the whole framework into three parts, namely, binary backbone, binary feature pyramid and binary heads.}
	\label{fig:detection}
%	\vspace{-1em}
\end{figure*}

We further generalize GroupNet to the object detection task. 
We work on the one-stage detector, which consists of backbone, feature pyramid and heads.
We directly use the backbone network pretrained on the ImageNet classification task to initialize the detection backbone. For the feature pyramid, it attaches a $1\times1$ and $3\times3$ convolutional layer at each resolution to adapt the feature maps. Since it lacks of structural information like the backbone network, we therefore apply the layer-wise binary decomposition in Section~\ref{sec:layerwise}. 
Furthermore, the detection heads occupy a large portion of complexity in the whole detection framework. And each head is comprised of several consequent layers which is similar to a basic residual block. To preserve the structural information, following the spirit of group-wise binary decomposition strategy in Section~\ref{sec:groupwise}, we propose to approximate each head as a whole. The structure is illustrated in Figure~\ref{fig:detection}.

We note that except the last layers for classification, center-ness and regression, other parameters of the heads are not shared 
across all feature pyramid levels which is the opposite of the full-precision counterpart. 
Such a design is motivated by the observation in \cite{chen2021aqd} that there is a large divergence of activation distributions across different feature pyramid levels. To capture accurate batch statistics, we privatize BN layers
for each pyramid level. Moreover, as a single binary convolutional layer has worse representational capability than the fixed-point one, we also use separate convolutional weights to ensure the model capacity.
More optimization details are explained in Section~\ref{exp:coco}.

\section{Discussions} \label{sec:discussion}

{\small\textbf{Differences from fixed-point quantization approaches.}}
Our GroupNet is different from fixed-point quantization approaches~\cite{zhang2018lq, zhuang2018towards, zhou2016dorefa, mishra2018apprentice} in both the quantization algorithm and its underlying inference implementation. 

Note that in conventional fixed-point methods, the inner product between fixed-point weights and activations can be computed by bitwise operations.
Let ${\bw} \in {\mathbb{R}^M}$ and $\ba \in {\mathbb{R}^M}$
be the weights and activations, respectively.  they can be encoded by a linear combination of binary bases, respectively. 
In particular, $\bw$ and $\ba$ can be encoded by ${\bf{b}}_i^w \in {\{ - 1,1\} ^M}$ and ${\bf{b}}_i^a \in {\{ - 1,1\} ^M}$, where $i = 1,...,P$, respectively. Let $Q_P(\cdot)$ be any quantization function, and for simplicity, we here consider uniform quantization only. Then, the inner product of $\bw$ and $\ba$ can be approximated by
\begin{equation} \label{eq:fixed-point}
{Q_P}({{\bf{w}}^\mathrm{T}}){Q_P}({\bf{a}}) = \sum\limits_{i = 0}^{P-1} {\sum\limits_{j = 0}^{P-1} {{2^{i + j}}} } ({\bf{b}}_i^w \xor {\bf{b}}_j^a).
\end{equation}
where $\xor$ indicates the binary inner product, which can be efficiently implemented by
$\rm popcount$ and $\rm xnor$ bit-wise instructions that are commonly equipped in modern computers. According to Eq. (\ref{eq:fixed-point}), the computational complexity is $O(MP^2)$ for the fixed-point inner product.
Bearing the group-wise binary decomposition in Section \ref{sec:decomposition} in mind, it can be easily realized that the proposed GroupNet with $K$ bases has the same computational complexity with the conventional $P$-bit fixed point quantization when $K = 2^P$. However, compared with the $P$-bit fixed point quantization, the proposed GroupNet with $2^P$ bases introduces extra additions from the fusion of decomposition branches. It is worth noting that the convolution is implemented with two steps: $\rm im2col$ and $\rm GEMM$. The high-precision additions can be efficiently merged into the $\rm im2col$ operation of the succeeding layer during implementation. For $\rm im2col$ operation, most of the time is spend on data rearrangement, an extra tensor addition will have a very limited influence on its execution time.
Moreover, the extra additions only account for a small portion of the overall network complexity, which intrinsically limits the impact on the overall speed.
To justify our analysis, we provide speedup evaluations in Section \ref{sec:speed_benchmarks}. We also provide the analysis of the runtime training and inference memory cost in Section \ref{exp:runtime_memory}.

To emphasize, 
the proposed GroupNet enables much more flexible design of the quantization algorithm. 
Benefiting from the group-wise binary decomposition, the proposed GroupNet allows more fine-grained exploration space, where the quantization levels can be chosen from the continuous positive integer domain. In contrast, conventional fixed-point quantization requires the quantization levels to be the power of 2. For example, the bit-width of GroupNet with 5 bases is between 2-bit and 3-bit. Thus, GroupNet enjoys a flexible trade-off between complexity and accuracy by setting appropriate $K$.
Moreover, as introduced in Section \ref{sec:segmentation} and Section \ref{sec:detection}, 
our GroupNet is demonstrated to be more efficient in exploiting task-specific information or structures to compensate the quantization information loss.

Moreover, we define the model size of the full-precision model as $\Omega$. Then, the model size for GroupNet with $K$ bases becomes $K\Omega/32$ and the one for $P$-bit fixed-point quantization is $P\Omega/32$.

Based on the above analysis, we summarize that the proposed GroupNet
introduces small additional run-time cost caused by the extra addition operations, however, provides benefits in algorithm flexibility and performance.

\noindent\textbf{\kui{Differences between GroupNet
and 
other multiple binarizations methods.}}
In \cite{lin2017towards}, a linear combination of binary weight/activations bases are obtained from the full-precision weights or activations without being directly learned. In contrast, we directly design the binary network structure, where binary weights are end-to-end optimized. ~\cite{fromm2018heterogeneous, tang2017train, guo2017network, li2017performance} propose to recursively approximate the residual error and obtain a series of binary maps corresponding to different quantization scales.
However, it is a sequential process which cannot be paralleled. Also, all previous multiple binarizations methods belong to local tensor approximation. 
In contrast to value approximation, we propose a group decomposition approach to approximate the full-precision network. Moreover, tensor-based methods are tightly designed to local value approximation and may be hardly generalized to other tasks accordingly. In addition, our group decomposition strategy achieves much better performance than tensor-level approximation as shown in Section~\ref{sec:group_space}.

\section{Experiments}

We define several methods for comparison as follows:
\noindent\textbf{Group\-Net-A:} It implements the full model with learnt soft connections described in Section~\ref{sec:learn}. Following Bi-Real Net~\cite{liu2020bi}, we apply skip connection bypassing every binary convolution to improve the convergence. Note that due to the binary convolution, skip connections are high-precision which can be efficiently implemented using fixed-point addition.
\textbf{Gr\-ou\-p\-N\-e\-t-B:} Based on GroupNet-A, we keep the $1 \times 1$ downsampling skip connections to high-precision (\ie, 8-bit) similar to~\cite{bethge2018training, Liu_2018_ECCV}.
\noindent\textbf{GroupNet-C:} Based on G\-r\-ou\-p\-N\-e\-t-B, we replace the soft routing mechanism with the dynamic hard gating. By default, we train 8 bases and dynamically select 4 bases during inference.

\subsection{Evaluation on ImageNet} \label{sec:imagenet}

\subsubsection{Experimental settings}

\begin{table*}[!tb]
\caption{Performance comparisons of different binarization methods on ImageNet. We use operations count (OPs) to measure the computational cost. ``-'' denotes that the results are not reported. ``W'' and ``A'' refer to the weight and activation bitwidth respectively. We use the customized MobileNetV1 structure following ReActNet. 
}
\centering
\scalebox{0.9}
{
\begin{tabular}{|c|cccccc|}
\hline
Network & Method & Bitwidth (W/A) &\#Parameters (Mbit) & OPs ($\times 10^8$) & Top-1 Acc. (\%) & Top-5 Acc. (\%)   \\
\hline\hline
\multirow{12}{*}{ResNet-18} & Full-precision & 32/32 &374.0  &18.17  & 69.7 &89.4 \\
\cdashline{2-7}
&BNNs \cite{hubara2016binarized} &1/1  &27.2  &1.47  &42.2 &67.1 \\
&XNOR-Net \cite{rastegari2016xnor} &1/1 &33.7  &1.67  &51.2  &73.2  \\
&Bi-RealNet \cite{Liu_2018_ECCV} &1/1  &33.6 &1.63 &56.4 &79.5 \\
&Real-to-Binary Net \cite{martinez2020training} &1/1 &44.6  &1.83  &65.4  &86.2 \\
&HCE \cite{bulat2021high} &1/1 &293.7 &1.37 &67.5 &87.5 \\
\cdashline{2-7}
&BitSplit \cite{kim2020algorithm}  &(1/1) $\times$ 4 &-  &2.47  & 58.4 &80.6 \\ 
&BENN \cite{zhu2019binary} & (1/1) $\times$ 6 &168.9  &8.70  &61.0  &-   \\
&Circulant CNN \cite{liu2019circulant} & (1/1) $\times$ 4 &-  &2.47  &61.4 &82.8 \\
\cdashline{2-7}
&GroupNet-A &(1/1) $\times$ 4 &50.0  &2.25  &65.2  &85.7  \\ 
&GroupNet-B &(1/1) $\times$ 4 &54.8  &2.43  &67.3 &87.6  \\
&GroupNet-C &(1/1) $\times$ 4 &105.5  &2.43  &\bf{68.2}  &\bf{88.3} \\
\hline\hline
\multirow{7}{*}{ResNet-34} & Full-precision & 32/32 &697.5  &36.68 &73.2  &91.4   \\
\cdashline{2-7}
&Bi-RealNet \cite{Liu_2018_ECCV} &1/1 &43.8 &1.93  &62.2  &83.9 \\
&BENN \cite{zhu2019binary} &(1/1) $\times$ 6 &230.9  &10.44  &64.7  &84.4 \\
&GroupNet-A &(1/1) $\times$ 4 &103.7  &3.40  &68.7  &88.3  \\ 
&GroupNet-B &(1/1) $\times$ 4 &108.5  &3.58  &70.7  &89.6  \\
&GroupNet-C &(1/1) $\times$ 4 &188.2  &3.58  &\bf{72.2}  &\bf{90.5} \\
\hline\hline
\multirow{7}{*}{ResNet-50} & Full-precision & 32/32 &817.8  &41.00  &76.0 &92.9 \\
\cdashline{2-7}
&Bi-RealNet \cite{Liu_2018_ECCV} &1/1 &176.8  &5.36  &62.6  &83.9  \\
&BENN \cite{zhu2019binary} &(1/1) $\times$ 6 &545.9  &10.93  &66.2  &85.8  \\
&GroupNet-A &(1/1) $\times$ 4 &117.0  &3.68  &69.8  &88.0  \\ 
&GroupNet-B &(1/1) $\times$ 4 &194.6  &7.05  &71.3  &90.2 \\
&GroupNet-C &(1/1) $\times$ 4 &217.6  &7.05  &\bf{73.4}  &\bf{91.0} \\
\hline\hline
\multirow{7}{*}{MobileNetV1$^{}$} & Full-precision & 32/32 &937.2 &48.33 &72.4  &-   \\
\cdashline{2-7}
&Bi-RealNet \cite{Liu_2018_ECCV} &1/1 &72.5  &1.90  &58.2  &81.0  \\
&ReActNet-C$^{*}$ \cite{liu2020reactnet} &1/1 &83.4 &2.14 &70.0  &- \\
&BENN \cite{zhu2019binary} &(1/1) $\times$ 6 &370.5  &5.32  &63.0  &84.0  \\
&GroupNet-A &(1/1) $\times$ 4 &124.0   &3.19  &66.8  &86.2  \\ 
&GroupNet-B &(1/1) $\times$ 4 &133.8  &4.16   &70.8  &89.5  \\
&GroupNet-C &(1/1) $\times$ 4 &259.3  &4.16   &\bf{72.1}  &\bf{90.3}  \\
\hline\hline

\end{tabular}
}
\begin{tablenotes}
     \item \footnotesize $^{*}$ denotes that we remove the knowledge distillation loss for fair comparison.
\end{tablenotes}

\label{tab:binary_compare}
\end{table*}

\noindent\textbf{Dataset.} The proposed method is first evaluated on ImageNet (ILSVRC2012) \cite{russakovsky2015imagenet} dataset. ImageNet is a large-scale dataset which has $\sim$1.2M training images from 1K categories and 50K validation images. Several representative networks are tested: ResNet-18 \cite{he2016deep}, ResNet-34, ResNet-50 and MobileNetV1 \cite{howard2017mobilenets}.
As discussed in Section~\ref{sec:discussion},
we compare the proposed approach with binary neural networks in Table~\ref{tab:binary_compare} and fixed-point approaches in Table~\ref{tab:groupnet_vs_fixedpoint}, respectively.

\noindent\textbf{Implementation details.}
As in \cite{rastegari2016xnor, Cai_2017_CVPR, zhou2016dorefa, zhuang2018towards}, we binarize the weights and activations of all convolutional layers, except that the first and the last layers are quantized to 8-bit. In all ImageNet experiments, training images are resized to $256 \times 256$, and a $224 \times 224$ crop is randomly sampled from an image or its horizontal flip, with the per-pixel mean subtracted. We do not use any further data augmentation in our implementation. We use a simple single-crop testing for standard evaluation. No bias term is utilized.
Following \cite{martinez2020training}, training is divided into two stages. In the first stage, we train a network with binary activations and real-valued weights from scratch. In the second stage, we inherit the weights from the first step as the initial value and fine-tune the network with weights and activations both being binary. For both stages, we use the Adam optimizer with a linear learning rate decay scheduler, an initial learning rate of 5e-4 and a batch size of 256. We optimize the network for maximum 90 epochs in each stage. The weight decay is set to 1e-5 for the first stage and 0 for the second stage.
Following~\cite{Cai_2017_CVPR, zhuang2018towards}, no dropout is used due to binarization itself can be treated as a regularization. We initialize the dynamic hard gating parameters $\boldsymbol{\nu}$ using Kaiming initialization \cite{he2015delving} and optimize them in conjunction with other network parameters using backpropagation.
We apply layer-reordering to the networks as: $\rm Sign$ $\to$ $\rm Conv$ $\to$ $\rm PReLU$ $\to$ $\rm BN$. Inserting $\rm PReLU(\cdot)$ after convolution is important for convergence. We follow the BN folding literature \cite{jacob2017quantization, chen2021aqd} to fuse BN parameters into convolutional weights during inference. During training, the BN parameters and the convolutional weights are updated separately.
Our simulation implementation is based on PyTorch \cite{paszke2017automatic}.

\noindent\textbf{OPs calculation.} We follow the calculation method in \cite{martinez2020training}, where we count the binary operations (BOPs) and floating point operations (FLOPs) separately. Then the total operations (OPs) is calculated by OPs = BOPs/64 + FLOPs, following the literature \cite{liu2020reactnet, Liu_2018_ECCV, rastegari2016xnor} in BNNs.

\noindent\textbf{Model size calculation.} Following the common practise in the literature \cite{rastegari2016xnor, Liu_2018_ECCV}, the model size is calculated by $32 \times N_f + P \times N_P + N_b$, where $N_f$, $N_P$ and $N_b$ are the number of full-precision, $P$-bit fixed-point and binary weights, respectively.

\subsubsection{Comparison with binary neural networks} \label{sec:compare_binary}

Since we employ binary weights and binary activations, we directly compare to the previous state-of-the-art binary approaches, including standard binary neural networks BNN~\cite{hubara2016binarized}, XNOR-Net~\cite{rastegari2016xnor}, Bi-RealNet~\cite{Liu_2018_ECCV}, ReActNet \cite{liu2020reactnet}, Real-to-Binary Net \cite{martinez2020training}, HCE \cite{bulat2021high}; and multiple binarization methods 
BitSplit \cite{kim2020algorithm}, Circulant CNN \cite{liu2019circulant}, BENN \cite{zhu2019binary}. 
We report the results in Table~\ref{tab:binary_compare} and summarize the following points. 1): 
When comparing with other multiple binarizations methods, our GroupNet shows better performance with comparable complexity. 
In comparison to directly binarizing networks, GroupNet also achieves better performance with moderate complexity increase. 
In summary, our approach achieves an improved trade-off between computational complexity and prediction accuracy.
2): We observe that GroupNet-B outperforms GroupNet-A by a large margin. It justifies keeping $1 \times 1$ downsampling skip connections to high-precision is crucial for achieving high performance. 
3): For Bottleneck structure in ResNet-50, we find larger quantization error than the counterparts using basic blocks with $3 \times 3$ convolutions in ResNet-18 and ResNet-34. The similar observation is also found by~\cite{bethge2018learning}. We assume that this is mainly attributable to the $1 \times 1$ convolutions in Bottleneck. The reason is $1 \times 1$ filters are limited to two states only (either 1 or $-1$) and they have very limited learning power. Moreover,
the bottleneck structure reduces the number of filters significantly, which implies the gradient paths are greatly reduced, leading to increased optimization difficulty.

\subsubsection{Comparison with fixed-point approaches} \label{sec:fixed-point}

\begin{table*}[!tb]
\caption{Performance comparisons of different methods on ImageNet. We use operations count (OPs) to measure the computational cost. ``-'' denotes that the results are not reported. “W” and “A” refer to the weight and activation bitwidth respectively.
}
\centering
\scalebox{0.95}
{
\begin{tabular}{|c|cccccc|}
\hline
Network & Method & Bitwidth (W/A) &\#Parameters (Mbit) & OPs ($\times 10^8$) & Top-1 Acc. (\%) & Top-5 Acc. (\%)   \\
\hline\hline
\multirow{7}{*}{ResNet-18} & Full-precision & 32/32 &374.0 &18.17  &69.8  &89.1   \\
\cdashline{2-7}
&GroupNet-C  &(1/1) $\times$ 4 &105.5  &2.43  &\bf{68.2}  &\bf{88.3}  \\
&LSQ~\cite{esser2020learned}   &2/2  &61.6 &2.25  &67.6  &87.6  \\
&QIL~\cite{jung2019learning}   &2/2 &61.6  &2.25  &65.7  &-  \\ 
&PACT~\cite{choi2018pact}   &2/2  &61.6  &2.25  &64.4  &-  \\
&LQ-Net~\cite{zhang2018lq}  &2/2  &61.6  &2.25  &64.9  &85.9  \\
\hline\hline
\multirow{7}{*}{ResNet-34} & Full-precision & 32/32 &697.5 &36.68  &73.3  &91.4   \\
\cdashline{2-7}
&GroupNet-C   &(1/1) $\times$ 4  &188.2  &3.58  &\bf{72.2}  &\bf{90.5}  \\
&LSQ~\cite{esser2020learned}   &2/2 &102.3  &3.40   &71.6  &90.3  \\
&QIL~\cite{jung2019learning}   &2/2 &102.3  &3.40  &70.6  &-  \\ 
&PACT~\cite{choi2018pact}   &2/2  &102.3  &3.40  &-  &-  \\
&LQ-Net~\cite{zhang2018lq}  &2/2 &102.3 &3.40  &69.8   &89.1  \\
\hline

\end{tabular}
}
\label{tab:groupnet_vs_fixedpoint}
\end{table*}

\begin{table}
	\centering
	\caption{Performance of GroupNet and the width multiplier on ImageNet.}
	\scalebox{1.0}
	{
		\begin{tabular}{c | c c}

			Model &Top-1 \%  &Top-5 \% \\
			\hline
			GroupNet-A & \bf{65.2}  &\bf{85.7}   \\
			Width multiplier &63.5  &85.2  \\
			
	\end{tabular}}
	\label{tab:classification_width}
\end{table}

\begin{table}
	\centering
	\caption{Comparisons between several group-wise decomposition strategies. Top-1 and Top-5 accuracy (\%) gaps to the corresponding full-precision ResNet-18 network are also reported.}
	\scalebox{0.85}
	{
		\begin{tabular}{c| c| c c c c}
			Model &Bases &Top-1 &Top-5 &Top-1 gap &Top-5 gap \\\hline
			Full-precision &1 &69.7 &89.4 &- &-  \\ 
			GroupNet-A &4  &\bf{65.2} &\bf{85.7}  &\bf{4.5} &\bf{3.7}  \\
			GBD v1 & 4 &63.7  &85.2  &6.0  &4.2   \\
		    GBD v2 & 4 &62.9 &84.7  &6.8  &4.7  \\
			GBD v3 & 4 &59.8 &82.7 &9.9  &6.7  \\
			LBD &4 &57.8 &80.1 &11.9  &9.3 \\
	\end{tabular}}
	\label{tab:groupspace}
\end{table}

Since we use $K$ binary group bases, we compare our approach with at least $\sqrt K$-bit fixed-point approaches. In Table~\ref{tab:groupnet_vs_fixedpoint}, we compare our approach with the state-of-the-art fixed-point approaches LSQ \cite{esser2020learned}, QIL~\cite{jung2019learning}, PACT~\cite{choi2018pact} and LQ-Nets~\cite{zhang2018lq}. 
 
All the comparison results are directly cited from the corresponding papers.
From the results, we observe that with comparable computational complexity, the proposed GroupNet surpasses the state-of-the-art fixed-point methods, which demonstrates the flexibility and effectiveness of our method.

\subsubsection{Extension on MobileNetV1}

We further propose to apply GroupNet based on ReActNet. Specifically, ReActNet customizes the MobileNetV1 structure as the binarization backbone and GroupNet-C treats the ReActNet as the base structure. Following ReActNet, we make three modifications. 1): The $3 \times 3$ depth-wise and the $1 \times 1$ point-wise convolutional blocks in the MobileNetV1 are replaced by the $3 \times 3$ and $1 \times 1$ vanilla convolutions in parallel with shortcuts respectively. 2): For the reduction block, 
we duplicate the input activation and concatenate two blocks with the same inputs to address the channel number difference, permitting to add identity skip connections.
3): We also replace $\rm Sign$ and $\rm ReLU$ to $\rm RSign$ and $\rm RPReLU$ following ReActNet, respectively.

The empirical results show that our GroupNet is complementary to high-performance binary architectures ($K=1$) by exploring how to ensemble the binary bases ($K > 1$) effectively to approximate the original full-precision network. Based on an advanced binary base architecture, our GroupNet can achieve better performance.

\subsection{Ablation study on ImageNet classification}

\subsubsection{Width multiplier vs. structured approximation} \label{sec:width_multiplier}

We further compare the ``structured approximation'' with the width multiplier baseline \cite{howard2017mobilenets}, which simply multiplies the channel number by a fixed ratio. To make the two settings directly comparable, we set $K=4$ and the width multiplier to 2. The results are shown in Table \ref{tab:classification_width}.

We observe that GroupNet outperforms the width multiplier baseline on ResNet-18. This further justifies the group-wise approximation can better preserve the information.

\subsubsection{Effect of the group space}
\label{sec:group_space}

To show the importance of the space design, we define more methods for comparison as follows:
\noindent\textbf{LBD:} It implements the layer-wise binary decomposition strategy described in Section~\ref{sec:layerwise}.
\noindent\textbf{GBD v1:} We implement with the group-wise binary decomposition strategy, where each base consists of one block. It corresponds to the approach described in Eq. (\ref{eq:2}) and is illustrated in Figure~\ref{fig:main} (d).
\noindent\textbf{GBD v2:} Similar to GBD v1, the only difference is that each group base has two blocks. It is illustrated in Figure~\ref{fig:main} (e) and is explained in Eq. (\ref{eq:3}).
\noindent\textbf{GBD v3:} It is an extreme case where each base is a whole network, which can be treated as an ensemble of a set of binary networks. \
We present the results in Table \ref{tab:groupspace}.

From the results, we have two main observations and analysis. 1): The group-wise decomposition strategies (GBD v1-v3, GroupNet-A) outperform the layer-wise counterpart (LBD) by a large margin. Our speculation is that skip connections have been demonstrated to play an important role in increasing representational capability and improving gradient backpropagation in BNNs literature \cite{Liu_2018_ECCV}. In this sense, group-wise variants have $K$ times more skip connections than LBD, which explains to some extent why the group-wise approximation performs much better. 2): As shown in product quantization \cite{jegou2011product}, a fine quantization granularity is theoretically guaranteed to reduce the approximation error. Therefore, approximating one block as a whole is able to achieve the best trade-off since it simultaneously keeps the most skip connections at the fine granularity. This is reflected by the empirical results that the block-wise approximation (\eg, GroupNet-A and GBD v1) achieves the highest accuracy.

\subsubsection{Effect of the number of bases} \label{sec:bases}
%\vspace{-2mm}
\begin{table}
	\centering
	\caption{Validation accuracy (\%) of Group-Net on ImageNet with different number of bases. All cases are based on the ResNet-18 network with binary weights and activations.}
	\scalebox{0.85}
	{
		\begin{tabular}{c| c| c c c c c}
			Model &Bases &Top-1 &Top-5 &Top-1 gap &Top-5 gap \\\hline
			Full-precision &1  &69.7 &89.4 &- &-  \\
			GroupNet-A &1  &58.1  &80.8  &11.6  &8.6   \\
			GroupNet-A &3  &63.8 &84.9 &5.9  &4.5  \\
			GroupNet-A &4  & 65.2  &85.7  &4.5 &3.7 \\
			GroupNet-A &5  & \bf{66.0}  &\bf{86.2}  &\bf{3.7}  & \bf{3.2}
	\end{tabular}}
	\label{tab:paper_bases}
\end{table}

We further explore the influence of number of bases $K$ to the final performance in Table~\ref{tab:paper_bases}. When the number is set to 1, it corresponds to directly binarize the original full-precision network and we observe apparent accuracy drop compared to its full-precision counterpart. With more bases employed, we can find the performance steadily increases. The reason can be attributed to the better approximation of the floating-point network, which is a trade-off between accuracy and complexity. It can be expected that with enough bases, the network should has the capacity to approximate the full-precision network precisely.

\subsubsection{Effect of the soft gate}

\begin{table*}[hbt!]
	\centering
	\caption{The effect of soft gates on ImageNet.}
	\scalebox{1.0}
	{
		\begin{tabular}{c| c | c c c}
			\multicolumn{2}{c|}{Model} &Full &GroupNet-A (w/o softgates) &GroupNet-A \\\hline
			\multirow{2}{*}{ResNet-18}& Top-1 \% &69.7  &64.1  &\bf{65.2} \\ 
			&Top-5 \% &89.4  &85.0  &\bf{85.7}  \\\hline
			\multirow{2}{*}{ResNet-34}& Top-1 \% &73.2  &66.4  &\bf{68.7} \\ 
            &Top-5 \% &91.4  &86.2   &\bf{88.3}    \\\hline
			\multirow{2}{*}{ResNet-50}& Top-1 \% &76.0  &67.7  &\bf{69.8} \\ 
			&Top-5 \% &92.9  &86.7  &\bf{88.0}  \\
	\end{tabular}}
	\label{tab:soft_gate}
\end{table*}

In this section, we further analysis the effect of soft gate described in Section \ref{sec:soft_gating}. We show 
the quantitative results in Table~\ref{tab:soft_gate}.
From the results, we can observe consistent accuracy improvement for various architectures. This shows that increasing the gradient paths and learning the information flow within BNNs is important for maintaining the performance. For instance, on ResNet-34, learning the soft gates can improve the Top-1 accuracy by $2.1\%$.

\subsection{Evaluation on semantic segmentation} \label{exp:pascal}

\noindent In this section, we further evaluate GroupNet on the PASCAL VOC 2012 semantic segmentation benchmark \cite{everingham2010pascal} which contains 20 foreground object classes and one background class. 
The original dataset contains 1,464 (\emph{train}), 1,449 (\emph{val}) and 1,456 (\emph{test}) images. 
The dataset is augmented by the extra annotations from~\cite{hariharan2011semantic}, resulting in 10,582 training images. The performance is measured in terms of averaged pixel intersection-over-union (mIOU) over 21 classes.

\noindent\textbf{Implementation details.} Our experiments are based on FCN~\cite{long2015fully}, where we adjust the dilation rates of the last two stages in ResNet with atrous convolution to make the output stride equal to 8. We empirically set dilation rates to be (4, 8) in last two stages. Similar to the structure of FCN-32s and FCN-16s, we define our modified baselines as FCN-8s-C5 and FCN-8s-C4C5, where C4 and C5 denote extracting features from the final convolutional layer of the 4-th and 5-th stage, respectively. We first pretrain the binary backbone on ImageNet dataset and fine-tune it on PASCAL VOC. During fine-tuning, we use Adam with initial learning rate=1e-4, weight decay=0 and batch size=16. We set the number of bases $K=4$ in experiments. We train 40 epochs in total and decay the learning rate by a factor of 10 at 20 and 30 epochs. We do not add any auxiliary loss and ASPP.

\subsubsection{Experiments on FCN}

\begin{table}[t!]
	\centering
	\caption{Performance of GroupNet on the PASCAL VOC 2012 validation set with FCN.}
	\scalebox{0.95}
	{
		\begin{tabular}{c| c |c c}
			Backbone &Model &mIOU & $\Delta$\\
			\hline
			\multirow{3}{*}{ResNet-18, FCN-8s-C5}& Full-precision &64.9 &- \\
			&LSQ (2-bit) &60.8  &4.1  \\
			&GroupNet-C &\bf{61.5} &\bf{3.4}  \\\hline
			
			\multirow{3}{*}{ResNet-18, FCN-8s-C4C5}& Full-precision & 67.3 &-\\ 
			&LSQ (2-bit)  & 62.9  & 4.4   \\
			&GroupNet-C  & \bf{63.6}  & \bf{3.7}   \\\hline
			
			\multirow{3}{*}{ResNet-34, FCN-8s-C5}& Full-precision &72.7 &- \\ 
			&LSQ (2-bit) &68.5   &4.2  \\
			&GroupNet-C  & \bf{69.3}  &\bf{3.4}  \\ \hline
			
			\multirow{3}{*}{ResNet-50, FCN-8s-C5}
			& Full-precision &73.1  &-\\ 
			&LSQ (2-bit)  &69.6   &3.5  \\
			&GroupNet-C & \bf{70.0}  &\bf{3.1}   \\
			
	\end{tabular}}
	\label{tab:segmentation}
\end{table}

\begin{table}[hbt!]
	\centering
	\caption{Performance of GroupNet-C with BPAC on PASCAL VOC 2012 validation set with FCN.}
	\scalebox{0.8}
	{
		\begin{tabular}{c | c |c}
			Backbone &Model &mIOU \\
			\hline
			\multirow{3}{*}{ResNet-18, FCN-8s-C5}
			& Full-precision (multi-dilations) &67.6 \\
			& LSQ (2-bit) & 64.0 \\
			&GroupNet-C + BPAC  & \bf{66.2}  \\ \hline
			
			\multirow{3}{*}{ResNet-18, FCN-8s-C4C5}
			& Full-precision (multi-dilations)  &70.1 \\ 
			& LSQ (2-bit) & 66.5 \\
			&GroupNet-C + BPAC & \bf{69.0}  \\\hline	
			
			\multirow{3}{*}{ResNet-34, FCN-8s-C5}
			& Full-precision (multi-dilations) &75.0  \\ 
			& LSQ (2-bit) & 71.3 \\
			&GroupNet-C + BPAC & \bf{73.9} \\\hline	
			
			\multirow{3}{*}{ResNet-50, FCN-8s-C5}
			& Full-precision (multi-dilations) &75.5 \\ 
		    & LSQ (2-bit) & 72.6 \\
			&GroupNet-C + BPAC & \bf{74.4}  \\
	\end{tabular}}
	\label{tab:segmentation_BPAC}
\end{table}

The main results are reported in Table~\ref{tab:segmentation} and Table~\ref{tab:segmentation_BPAC}. From the results in Table~\ref{tab:segmentation}, we can observe that when all bases using the same dilation rate, there is an obvious performance gap with the full-precision counterpart. This performance drop is consistent with the classification results on ImageNet dataset in Table~\ref{tab:groupnet_vs_fixedpoint}. It proves that the quality of extracted features has a great impact on the segmentation performance. Moreover, we also quantize the network to 2-bit using the state-of-the-art fixed-point quantization method LSQ. Compared with LSQ, we achieve better performance with comparable computational cost.

To reduce performance loss, we further employ diverse dilated rates on parallel binary bases to capture the multi-scale information without increasing any computational complexity. This formulates our final approach GroupNet-C + BPAC in Table~\ref{tab:segmentation_BPAC}, which shows significant improvement over the GroupNet-C counterpart. Moreover, the performance of G\-r\-ou\-p\-N\-e\-t-C + B\-P\-A\-C outperforms the full-precision baseline in Table~\ref{tab:segmentation}, which strongly justifies the flexibility of GroupNet.

\subsubsection{Experiments on DeepLab\-v\-3}
For training the Deep\-L\-a\-b\-v\-3 \cite{chen2017rethinking} baseline, we use Adam as the optimizer. The initial learning rate for training backbone network is set to 1e-4, and is multiplied by 10 for ASPP module. Similar to image classification, we keep the first layer and last classification layer to 8-bit. We employ the layer-wise approximation in quantizing the ASPP module. We set $K=4$ for both backbone and ASPP. The training details are the same with those in FCN except that we use the polynomial decay of learning rate. The results are provided in Table~\ref{tab:deeplab}.

Moreover, a problem still exists in training binary DeepLabv3. The ASPP module uses large dilation rates for the three $3\times3$ convolutions with $rates=\{12, 24, 36\}$ when \emph{output\_stride=8}. For training BNNs with Eq.~(\ref{eq:binary_activations}), we apply one-paddings to constrain activations to $\{-1, 1\}$. However, for atrous convolution with large rates, padding ones can introduce high bias and make the optimization difficult. To solve this problem, we instead binarize the activations to $\{0, 1\}$ following the quantizer in \cite{zhou2016dorefa}. Note that the numerical difference is only a scalar whether activations are represented by $\{-1, 1\}$ or $\{0, 1\}$ in bitwise operations to accelerate the dot products~\cite{zhou2016dorefa, zhang2018lq}. The importance for binarizing activations to $\{0, 1\}$ is shown in Table~\ref{tab:deeplab_ablation}.

\begin{table}[t!]
	\centering
	\caption{The difference for binarizing ASPP with $\{-1, 1\}$ and $\{0, 1\}$. The metric is mIOU.}
	\scalebox{1.0}
	{
		\begin{tabular}{c | c c c}

			Model &full-precision &$\{0, 1\}$  & $\{-1, 1\}$ \\
			\hline
			ResNet-18 &72.1 &64.0 &51.6 
	\end{tabular}}
%	\vspace{-1em}
	\label{tab:deeplab_ablation}
\end{table}

It is worth noting that the comparable counterpart of Deep\-Labv3 is the GroupNet-C + B\-P\-A\-C with FCN-8s-C5 in Table~\ref{tab:segmentation_BPAC}. The main difference is that in the proposed BPAC module, we directly incorporate the multiple dilation rates in the backbone network to capture multi-scale context. In contrast, DeepLabv3 embeds the ASPP module on top of the backbone network.
With the same \emph{output\_stride}, the computational complexity of FCN is lower than Deep\-Lab\-v3 since no additional ASPP is needed. With the simpler quantized FCN framework with BPAC, we can achieve comparable or even better performance than quantized Deep\-L\-a\-b\-v\-3. For example, with the ResNet-34 backbone, Group\-N\-e\-t-C + B\-P\-AC outperforms DeepLab\-v3 by 5.9 w.r.t.\  mIOU. It also shows that binarization on ASPP is sensitive to the final performance, since the binarization process constrains the feature magnitude to $\{0, 1\}$ which causes the multi-scale information loss.

\begin{table}[tb!]
	\centering
	\caption{Performance on the PASCAL VOC 2012 validation set with DeepLabv3.}
	\scalebox{1.0}
	{
		\begin{tabular}{c| c |c c}

			Backbone &Model &mIOU & $\Delta$\\
			\hline
			\multirow{3}{*}{ResNet-18} & Full-precision &72.1 &- \\
			&Backbone  &\bf{68.6}  &\bf{3.5} \\
			&Backbone + ASPP &65.0  &7.1   \\\hline
			\multirow{3}{*}{ResNet-34} & Full-precision &74.4 &- \\
			&Backbone  &\bf{71.9}  &\bf{2.5} \\
			&Backbone + ASPP & 68.0  &6.4 \\\hline
			\multirow{3}{*}{ResNet-50} & Full-precision &76.9 &- \\
			&Backbone &\bf{73.6}  &\bf{3.3} \\
			&Backbone + ASPP &70.2 & 6.7\\
			
	\end{tabular}}
%	\vspace{-1em}
	\label{tab:deeplab}
\end{table}

\subsection{Evaluation on object detection} \label{exp:coco}
In this section, we evaluate GroupNet on the general object detection task. Our experiments are conducted on the large-scale detection benchmark COCO~\cite{lin2014microsoft}. Following~\cite{lin2017feature, lin2017focal}, we use the COCO \textit{trainval35k} split (115K images) for training and \textit{minival} split (5K images) for validation. We also report our results on the \textit{test\_dev} split (20K images) by uploading our detection results to the evaluation server. Our experiments are based on FCOS \cite{tian2019fcos} and RetinaNet \cite{lin2017focal}.
In all settings, we set $K=4$.

\noindent\textbf{Implementation details.} 
In specific, the backbone is initialized by the pretrained weights on ImageNet classification. GroupNet is then fine-tuned with Adam with the initial learning rate of  5e-4 and the batch size of 16 for 90,000 iterations. The learning rate is decayed by 10 at iteration 60,000 and 80,000, respectively. Note that we keep updating the BN layers rather than fix them during training. Other hyper-parameters are kept the same with \cite{tian2019fcos, lin2017focal}.

\subsubsection{Performance evaluation}

\begin{table*}[t]
	\centering
	\caption{Performance on COCO validation set with different binarized components.}
		\scalebox{1.0}{
	\begin{tabular}{ c |c c c | c c c}
	
	 Model &AP &AP$_{50}$  &AP$_{75}$   &AP$_{S}$  &AP$_{M}$  &AP$_{L}$ \\\hline
	Full-precision&33.8 &51.8 &36.1 &19.3 &36.4 &44.4  \\
		Backbone &33.7  &51.6  &36.2   &19.0  &36.3  &44.5    \\
		Backbone + Pyramid &32.9  &50.3  &35.7  &18.8  &35.1 &42.8  \\
		Backbone + Pyramid + Heads (shared) &19.3  &36.6  &18.4 &10.5  &23.0   &28.8  \\
		Backbone + Pyramid + Heads (w/o shared) &30.1 &47.5  &33.2  &16.0  &31.2  &39.2  \\
	\end{tabular}}
	\label{tab:detection_component}
\end{table*}

We report the performance on FCOS in Tables~\ref{tab:detection_val} and \ref{tab:detection_test}, as well as RetinaNet in Table \ref{tab:retinanet}, respectively. We can observe that GroupNet achieves promising results over all ResNet architectures. For instance, with the ResNet-18 backbone, the gap of AP is only 3.7 on FCOS while we save considerable computational complexity. 
This strongly shows that the proposed GroupNet is a general approach that can be extended on many fundamental computer vision tasks.
Furthermore, we highlight that we are among the pioneering approaches to explore binary neural networks on object detection in the literature.

We compare with the state-of-the-art quantized object detector FQN \cite{li2019fully}, and also quantize the detector with LQ-Net using the proposed training strategy. We observe that GroupNet-C outperforms the two comparing methods. It shows that learning independent heads is more effective than freezing batch normalization layers in FQN to stabilize the optimization. Moreover, the better low-precision feature quality of GroupNet also contributes to superiority of the performance.

\begin{table*}[t]
	\centering
	\caption{Performance on the COCO validation set based on FCOS.}
	\scalebox{0.9}{
	\begin{tabular}{c |c | c | c | c c c | c c c}
		Backbone & Model & \#Parameters (Mbit) & FLOPs ($\times10^{10}$)  &AP &AP$_{50}$  &AP$_{75}$   &AP$_{S}$  &AP$_{M}$  &AP$_{L}$ \\\hline
		\multirow{4}{*}{ResNet-18}& Full-precision  &617.4 &15.5 &33.8 &51.8 &36.1  &19.3 &36.4 &44.4  \\
	    &FQN &86.2 &1.01  &26.2  &43.5 &26.7  &13.3  &29.5  &35.7  \\
	    &LQ-Net &86.2 &1.01 &28.0  &45.0  & 30.6 &15.0 &29.8 &36.6 \\
		&GroupNet-C &165.0 &1.03 &\bf{30.1} &\bf{47.5} &\bf{33.2} &\bf{16.0} &\bf{31.2} &\bf{39.2}  \\\hline
		\multirow{4}{*}{ResNet-34}& Full-precision &940.8 &19.3 &37.5 &55.9 &40.3 &22.6 &40.8 &47.4  \\
		&FQN &124.5 &1.25 & 28.8 &46.3  &30.0  &14.8 &31.0 &38.8  \\
	    &LQ-Net &124.5 &1.25 &30.8  &47.8 &33.4 &16.2  &32.5 & 40.0\\
		&GroupNet-C &245.8 &1.27  &\bf{32.8} &\bf{49.8}  &\bf{35.6}  &\bf{17.4}  &\bf{34.0}  &\bf{41.8}   \\ \hline
		\multirow{4}{*}{ResNet-50} &Full-precision &1034.0 &20.4  &38.6 &57.4 &41.4 &22.3 &42.5 &49.8  \\ 
		&FQN  &137.2 &1.32  &29.7  &47.1 &30.8  &15.3  &31.8 &39.3 \\
	    &LQ-Net  &137.2 &1.32  &32.1  &49.6 &34.8  &17.6 &33.2  &40.6  \\
		&GroupNet-C &260.4 &1.64  &\bf{33.9} &\bf{51.2} &\bf{37.3} &\bf{18.3} &\bf{35.0} &\bf{42.5}    \\
	\end{tabular}}
	\label{tab:detection_val}
\end{table*}

\begin{table*}[t]
	\centering
	\caption{Performance on the COCO test set based on FCOS.}
	\scalebox{0.9}{
	\begin{tabular}{c |c |c c c | c c c}
		Backbone & Model &AP &AP$_{50}$  &AP$_{75}$ &AP$_{S}$  &AP$_{M}$  &AP$_{L}$ \\\hline
		\multirow{4}{*}{ResNet-18}& Full-precision &33.9 &51.9 &36.4 &19.4 &35.6 &42.2  \\
		&FQN  &26.3 &43.8 &26.8 &13.5 &29.4 &35.4 \\
	    &LQ-Net  &28.2  &45.3  &30.7  &15.1 &29.6  &36.1  \\
		&GroupNet-C &\bf{30.2} &\bf{47.7}   &\bf{33.3}  &\bf{15.9}  &\bf{31.4}  &\bf{39.4}    \\\hline
		\multirow{4}{*}{ResNet-34}& Full-precision &37.8  &56.3  &41.0  &22.1  &40.1  &46.7  \\ 
		&FQN  &28.9 &46.6 &30.1 &14.5 &31.5 &37.8 \\
	    &LQ-Net  &30.9 &47.5 &33.5 &16.0 &32.3 &39.5  \\
		&GroupNet-C  &\bf{32.6}  & \bf{49.4} & \bf{35.7}  & \bf{17.3} &\bf{34.2} &\bf{41.7}   \\ \hline
		\multirow{4}{*}{ResNet-50}& Full-precision &38.8  &57.9  &41.9  &22.4  &41.5 &48.0 \\
		&FQN  &29.9  &47.4 &31.0  &14.9 &31.7 &39.0  \\
	    &LQ-Net  &32.0  &49.5 &35.1 &17.4  &32.7 &40.0 \\
		&GroupNet-C &\bf{33.8} &\bf{51.4} &\bf{37.5} &\bf{18.0} &\bf{35.1} &\bf{42.4}  \\
	\end{tabular}}
	\label{tab:detection_test}
\end{table*}

	\begin{table*}[t]
	\centering
	\caption{Performance on the COCO validation set based on RetinaNet.}
	\scalebox{0.9}{
	\begin{tabular}{c |c |c | c |c c c | c c c}
		Backbone & Model & \#Parameters (Mbit) & FLOPs ($\times10^{10}$) &AP &AP$_{50}$  &AP$_{75}$   &AP$_{S}$  &AP$_{M}$  &AP$_{L}$ \\\hline
		\multirow{3}{*}{ResNet-18}& Full-precision &685.8 &18.8  &32.9  &52.7  &34.8  &19.8  &35.6  &41.7   \\
	    &FQN (2-bit) &135.7 &1.43  &28.2  &46.5  &29.2  &15.6  &30.0  &37.7   \\
		&GroupNet-C &182.1 &1.44  &\bf{30.7} &\bf{48.0}  &\bf{32.8}  &\bf{16.4}  &\bf{32.2}  &\bf{39.8}  \\\hline
		\multirow{3}{*}{ResNet-34}& Full-precision &1008.6 &22.6  &36.2 &56.6 &38.7 &22.1  &39.6  &45.1  \\
		&FQN (2-bit) &175.8 &1.66  & 31.5  & 50.4   &33.2   &17.1  & 34.3  &41.4   \\
		&GroupNet-C  &262.4 &1.68  &\bf{33.4} & \bf{52.1} & \bf{36.4} & \bf{18.0}  & \bf{35.2}  & \bf{42.5}  \\ \hline
		\multirow{3}{*}{ResNet-50}& Full-precision &1215.6 &23.9 &37.8  & 58.5  & 40.7  & 22.8 &41.3   & 48.3  \\
		&FQN (2-bit) &203.1 &1.74 &31.9  &51.1 &33.5  &17.9  &35.1  &40.3 \\
		&GroupNet-C &305.7 &2.06  &\bf{34.2}  &\bf{52.5}  &\bf{37.5}  &\bf{18.9}  &\bf{35.6}  &\bf{43.3}   \\
	\end{tabular}}
	\label{tab:retinanet}
\end{table*}

\subsubsection{Detection components analysis}

We further analysis the affect of quantizing the backbone, feature pyramid and heads to the final performance, respectively. The results are reported in Table~\ref{tab:detection_component}.

From Table~\ref{tab:detection_component}, binarizing the backbone and the feature pyramid only downgrades the performance by a small margin. However, binarizing heads causes an obvious AP drop. It can be attributed to that heads are responsible for adapting the extracted multi-level features to the classification and regression objectives. As a result, its representability is crucial for robust detectors. However, the multi-level information is undermined when being constrained to $\{-1, 1\}$.
This shows that the detection modules other than the backbone are sensitive to quantization, and we leave it as our future work.
    
By comparing the results with respect to weight sharing, 
we observe the original sharing heads strategy that widely used in full-precision detection frameworks performs extremely bad in the binary setting. In specific, with ResNet-18 backbone, the AP gap between with and without weight sharing reaches by 10.8. It is worth noting that separating the parameters does not increase any additional computational complexity. Even though the number of parameters are increased (i.e., by $\sim4$ times in heads), the memory consumption is still significantly reduced due to the 1-bit storage.

\section{Acceleration on hardware} \label{sec:speedup}
%\vspace{-0.5em}

\begin{figure}
    \centering
    \caption{Example code to implement binary GEMM with GPU instructions. The following kernel will be executed in SIMD (single instruction, multiple data stream) manner on the GPU.}
    \label{fig:example_code}
    
    \begin{lstlisting}
__kernel void gemm_v1(
    __global half *dst,      // GEMM result buffer, in dimension of [M * N]
    __global uchar *src,     // activation, in dimension of [M * K]
    __global uchar *weight,  // weight parameters, in dimension of [N, K]
    __global half *bias,     // bias parameters
    int M, int N, int K,     
    half alpha               // quantization scale factor
    ) 
{
  int x = get_global_id(0);   // thread id 
  int y = get_global_id(1);   // thread id
  x = min(x, M-1);
  y = min(y, N-1);

  uchar buffer;
  uchar filter;
  short result;
  result = 0;

  short i;
  for(i=0; i<K; i++) {
    buffer = src[i*M + x];
    filter = weight[i*N + y];
    result += popcount(buffer ^ filter);  // obtain count of ones
  }
  result = (short)(K*8) - result * (short)2; // obtain binary inner product result
  half save;
  save = result * alpha;
  dst[y*M + x] = save + bias[y];
  return;
}
\end{lstlisting}

\end{figure}

\begin{table*}[htb!]
	\centering
	\caption{The execution time (us) of a single layer with different configurations.}
	\scalebox{0.9}{
	\begin{tabular}{c | c | c c c c c c c c c c c}
	    \hline
		Device & model &case1  &case2  &case3  &case4  &case5  &case6 &case7 &case8 &case9 &case10 &case11  \\\hline
		\multirow{3}{*}
		{Qualcomm 821} & 2-bit  & 1928  &3543 &7968 & 24914 & 93069 & 1512 & 2348 & 3546 & 7188 & 21187 & 75486 \\
		&GroupNet & 2014 & 3621 & 8163 & 25376 & 94550 & 1613 & 2487 & 3627 & 7311 & 21362 & 75982 \\ 
		\cdashline{2-13}
		& Relative &4.4\% &2.2\% &2.4\% &1.8\% &1.6\% &6.6\% &5.9\% &2.3\% &1.7\% &0.8\% &0.7\% \\ \hline
		\multirow{3}{*}{Qualcomm 835}  & 2-bit &2076 & 2969 & 6463 & 20528 & 73967 &1600   & 1900  & 2852 & 5646  & 15822 &69803  \\
		 &GroupNet   & 2255  & 3206 & 6930 & 21776 & 75010  & 1754 & 1998  & 2982  & 5746  & 17676 & 71140 \\
		 \cdashline{2-13}
		 & Relative &8.6\%  &8.0\% &7.2\% &6.1\% &1.4\% &9.6\% &5.1\% &4.6\% &1.8\% &11.7\%  &1.9\%  \\
		 \hline
		\multirow{3}{*}{Kirin 970}  & 2-bit   &900   &2234   & 2544  & 8040  & 31027 & 1048 & 1594 & 2415  & 2390  & 6750  & 24087  \\
		 &GroupNet & 1017 & 2603 & 2550 & 8030 & 31145 & 1060 & 1594 & 2535 & 2478 & 6764 & 24326  \\
		 \cdashline{2-13}
		 & Relative &13.0\% &16.5\% &0.2\% &-0.1\% &0.4\% &1.1\% &0.0\% &5.0\% &3.7\% &0.2\% &1.0\% \\
		 \hline
	\end{tabular}}
	\label{tab:2-bit_layerwise_speed}
\end{table*}

%\fi

\begin{table*}[hbt!]
\centering
\caption{Exact execution time ($\rm ms$) and speedup ratios for overall quantized layers. We run 5 times and report the results with mean and standard deviation.}
\label{tab:overall}
\scalebox{0.9}{
\begin{tabular}{c | c | c c c| c c c}
\hline 
 Device & Network & Binary & GroupNet-C & 2-bit & Binary vs. GroupNet-C & Binary vs. 2-bit & GroupNet-C vs. 2-bit \\
\hline
\multirow{2}{*} {Q835} & ResNet-18 & 12.1$\pm$0.2 &48.8$\pm$0.3  &44.8$\pm$0.2  &4.03 &3.70 &0.92   \\
  & ResNet-34 & 25.3$\pm$0.3 &98.0$\pm$0.5  &90.6$\pm$0.4  &3.87  &3.58  &0.93  \\
\hline 
\multirow{2}{*} {Q821} & ResNet-18 & 15.7$\pm$0.3 &55.4$\pm$1.6 & 52.5$\pm$1.0 &3.45  &3.34  &0.95  \\
  & ResNet-34 & 32.3$\pm$0.6 &110.3$\pm$2.6  &105.2$\pm$2.1  &3.41  &3.26  &0.95 \\
\hline 
\end{tabular}}
\end{table*}

%\iffalse
To investigate the real execution speed and memory cost of the proposed GroupNet and other related counterparts such as fixed-point quantization, we develop the acceleration code (GPU version only) on resource constrained platforms and provide the benchmark results.
Experimental platforms include HiSilicon Kirin 970, Qualcomm 821 as well as Qualcomm 835.
For fair comparison, we fix the frequency of the sub-systems (such as the CPU, GPU and DDR) if possible in order to prevent interference from the DVFS (dynamic voltage and frequency scaling) on the modern operating system. We fuse the batch normalization layers into the corresponding convolutional layers following \cite{jacob2017quantization, chen2021aqd}.
Multiple rounds of experiments are conducted and the profiling data is averaged for statistic stability.

\subsection{Execution speed benchmarks}
\label{sec:speed_benchmarks}

\noindent\textbf{Implementation details.}
We employ OpenCL (similar with CUDA and is a common programming language on the embedded platforms) to implement the GPU acceleration. Arithmetic and bit-wise logical operations, such as $\rm popcount$ and $\rm xor$ are supported in the OpenCL language. Dedicated software is developed for the proposed neural network acceleration.
As explained in Sec. \ref{sec:discussion}, the binary convolution is implemented with two steps: $\rm im2col$ and $\rm GEMM$. We provide the example implementation in Figure \ref{fig:example_code}.

\noindent\textbf{Results analysis.}
We first compare the layer-wise execution time between GroupNet (refer to Eq. (\ref{eq:1})) with 4 bases and 2-bit models.
%for a single layer.
The results are listed in Table \ref{tab:2-bit_layerwise_speed}. A total of 11 cases are included in this experiment. All cases are with convolutions of $3\times3$ kernel, padding 1 and stride 1. For convenience, we configure the input and output feature maps to have the same shape, where input channels/width/height = output channels/width/height, respectively. 
For the first five cases, we fix the channel number to be 64 and increase the resolution from 28 to 448 with a multiplier 2. For the last six cases, we fix the resolution to be $56 \times 56$ and double the channel number from 16 to 512.
From the results, we observe that the inference speed of GroupNet is slighter slower than that of the $2$-bit model due to the extra addition operations discussed in Section \ref{sec:discussion}. 
We set the batch size to 1 in all scenarios.
We also report the overall speed for all quantized layers in Table \ref{tab:overall} of BNNs, GroupNet-C (4 bases) and 2-bit models.
From Table \ref{tab:overall}, we first observe that the real execution time of GroupNet-C and 2-bit models is less than $4 \times$ slower than the binary case. 
Moreover, we also report the speedup of the 4 bases GroupNet-C against the $2$-bit model. Specifically, the relative speedup ranges from 0.92 to 0.95 across different architectures (ResNet-18 and ResNet-34) on various devices (Qualcomm 821 and 835). It implies the extra addition operations analyzed in Section \ref{sec:discussion} has small impact on the overall inference speed. In summary, the group-wise approximation is still hardware-friendly but more flexible and accurate.

We finally report the inference time of binary convolutions 
and high-precision additions respectively in Table \ref{tab:binary_conv_vs_add}, where we can observe that ``hAdd'' executes much faster than ``bConv''.

\begin{table*}[htb!]
	\centering
	\caption{Comparison on the inference time of the binary convolution and high-precision addition. `bConv' and `hAdd' indicate the binary convolution and the high-precision addition, respectively. 
	}
	\scalebox{0.9}{
	\begin{tabular}{c | c | c c c c c c c c c c c}
	    \hline
		Device & Mode & case1 & case2 & case3 & case4 & case5 & case6 & case7 & case8 & case9 & case10 & case11  \\\hline
		\multirow{2}{*}
		{Qualcomm 821} & bConv  & 588  & 1227 & 2634 & 8265 & 31106 & 476 & 737 & 1249 & 2259 & 6000 & 19914 \\
		& hAdd & 191 & 510 & 832 & 1769 & 5592 & 187 & 366 & 510 & 677 & 1038 & 1311 \\ 
		\hline
		\multirow{2}{*}{Qualcomm 835}  & bConv & 652 & 918 & 1915 & 6053 & 23718 & 526 & 692 & 946 & 1790 & 4839 & 17394 \\
		 & hAdd & 98 & 209 & 375 & 1051 & 3899 & 114 & 122 & 210 & 262  & 362 & 577 \\
		 \hline
		\multirow{2}{*}{Kirin 970}  & bConv & 337 & 742 & 849 & 2617 & 9762 & 317 & 465 & 698 & 693 & 1898 & 6292 \\
		 & hAdd & 73 & 113 & 272 & 957 & 3811 & 107 & 113 & 162 & 219 & 333 & 522 \\
		 \hline
	\end{tabular}}
	\label{tab:binary_conv_vs_add}
\end{table*}

\begin{table*}[htb!]
	\centering
	\caption{Memory consumption ($\rm MB$) with ResNet-18 for the proposed method and fixed-point quantization scheme. For memory during training, we set the training batch size to be 64 and measure the GPU memory cost. For storage needed in the inference mode, we report the consumption by the activations and weights, respectively, in a format of `a + w' (we do not include the memory consumption of the input layer and the output layer).}
	\scalebox{1.0} {
	\begin{tabular}{c | c c c | c c c c}
	    \multirow{3}{*}{Mode} & \multicolumn{3}{c|}{GroupNet-C} & \multicolumn{4}{c}{Fixed-point} \\
	    \cline{2-8}
	    & \multicolumn{3}{c|}{base} &  \multicolumn{4}{c}{a/w} \\
	    \cline{2-8}
	    & 2 & 3 & 4 & 1/1 & 2/2 & 4/4 & FP32/FP32 \\
	    \hline
	    training & 3953 & 5117 & 6275 & 4969 & 4969 & 4969 & -\\
	    \hline
	    inference & 2.06 + 4.59 & 2.06 + 6.89 & 2.06 + 9.18 & 1.43 + 2.30 & 1.65 + 4.59 & 2.11 + 9.18 & 9.70 + 73.44\\
	\end{tabular}
	}
	\label{tab:extra_memory}
\end{table*}

\subsection{Runtime memory consumption} \label{exp:runtime_memory}

We further explore the actual training and inference memory consumption of GroupNet-C and the fixed-point quantization scheme. Note that we focus more on the runtime memory footprint in practice because energy consumption is dominated by memory access \cite{han2016deep}. We measure the runtime memory consumption with ResNet-18 on ImageNet, as illustrated in Table \ref{tab:extra_memory}.

\noindent\textbf{Implementation details.}
When developing the acceleration code with OpenCL on the target platforms, we find the memory allocation is quite time consuming. Therefore, we leverage the compilation phase allocation strategy. Specifically, we build up the computation graph for the network inference and record the lifetime of each tensor. Only one static memory pool is allocated, where tensors are mapped into sub-buffers of this memory pool. When the lifetime of a tensor ends, the buffer is marked free and can be re-used by the following tensors. The mapping pattern is pre-determined at the compilation stage and hard-coded into the program, thus negligible time is introduced by the memory mapping during runtime.

\noindent\textbf{Results analysis.}
In the training phase, we use ``fake quantization'', where we simulate the quantization by rounding the tensors into binary ones, however still represent the data with floating-point. Each extra branch would introduce an extra memory cost (memory can not be re-used among different branches, as the tensors are saved for back-propagation). Thus, the training memory increases linearly with the number of bases $K$ in GroupNet. In contrast, the fixed-point solution has a fixed amount of memory consumption for all bit quantization training ($K=1$).

For the runtime memory footprint at the testing time, we count the activation storage and weight storage separately.
In particular, the activation storage of our GroupNet only brings a fixed amount of memory cost. It is because after the computation of each branch, the buffer is freed and re-used in the next branch. Specifically, a memory pool is allocated with the capacity being dependent by the largest buffer in the network. In practice, we found the largest buffer is occupied by the $\rm im2col$ result of the input convolutional layer (this layer has a large kernel size of $\rm 7 \times 7$) for ResNet.
However, for the fixed-point quantization scheme, the activation memory consumption increases with the bitwidth configuration. In other words, a fixed-point method needs to be pre-allocated memory for $P$-bit while GroupNet only needs a buffer for 1-bit (1 base). 
Note that accumulation is required for our GroupNet, which might not be freed when handling different branches. 
As a result, the preliminary size of the memory pool of our GroupNet ($2.06$ MB) is generally larger than that of the fixed-point solution ($ 1.43 $ MB for $1/1$).

Moreover, for weight parameters storage during inference, the memory consumption is directly decided by the base number $K$ or quantization bitwidth.

In summary, under comparable FLOPs, GroupNet-C (base $=4$) consumes higher runtime memory footprint than 2-bit fixed-point models, \ie, 11.24 MB vs. 6.24MB, where both are much less than the full-precision one, 83.14MB. The gap mainly comes from storing more parameters in our GroupNet.
\section{Conclusion}

\noindent In this paper, we have 
explored highly efficient and accurate CNN architectures with binary weights and activations. Specifically, we have proposed to directly decompose the full-precision network into multiple groups and each group is approximated using a set of binary bases which can be optimized in an end-to-end manner. We have also proposed to learn the decomposition automatically. To increase model capacity, we have introduced conditional computing to binary networks, where the bases in each group are dynamically executed.
Experimental results have proved the effectiveness of the proposed approach on the ImageNet classification task.
More importantly, we have generalized the proposed GroupNet
approach 
from image classification tasks to more challenging  fundamental computer vision tasks, namely
dense prediction tasks such as semantic segmentation and object detection.
We highlight that 
we may be among the first few approaches to apply binary neural networks on general
semantic segmentation and object detection tasks, and achieve 
encouraging 
performance on the PASCAL VOC and COCO datasets with binary networks.
Last,
we have developed the underlying acceleration code and speedup evaluation comparing with other quantization strategies is analyzed on several platforms, which serves as a strong benchmark for further research.
In the future, we will employ the latent-free optimizer \cite{helwegen2019latent} for BNNs that directly update the binary weights, to reduce the memory consumption during training. We will also develop acceleration code on X86, ARMs and FPGA platforms.

\section*{Acknowledgments}
M. Tan was partially supported by Key Realm R\&D Program of Guangzhou 202007030007.

% BibTeX users please use one of
%\bibliographystyle{spbasic}      % basic style, author-year citations
%\bibliographystyle{spbasic}      % basic style, author-year citations
\bibliographystyle{IEEEtran}
\bibliography{reference}
%\bibliographystyle{spmpsci}      % mathematics and physical sciences
%\bibliographystyle{spphys}       % APS-like style for physics
%\bibliography{}   % name your BibTeX data base

% % Non-BibTeX users please use
% \begin{thebibliography}{}
% %
% % and use \bibitem to create references. Consult the Instructions
% % for authors for reference list style.
% %
% \bibitem{RefJ}
% % Format for Journal Reference
% Author, Article title, Journal, Volume, page numbers (year)
% % Format for books
% \bibitem{RefB}
% Author, Book title, page numbers. Publisher, place (year)
% % etc
% \end{thebibliography}

\end{document}